\documentclass[preprint,10pt]{elsarticle}

\usepackage[T1]{fontenc}

\usepackage{graphicx}
\usepackage{hyperref}
\usepackage{lineno}   
\usepackage{parskip}
\usepackage{array} 
\usepackage{geometry}
\usepackage{float}
\usepackage{xcolor}
\modulolinenumbers[5]
\usepackage{amsmath}
\usepackage{cleveref}
\usepackage{amsfonts}
\usepackage{amssymb}
\usepackage{caption}
\usepackage{makecell}
\journal{ISPRS Journal of Photogrammetry and Remote Sensing}

\begin{document}

\begin{frontmatter}

\title{Toward Seasonal Guidelines for Robust Deep-Learning Sentinel-2 Building Detection  in Different Area Types}

\author[iitis]{Michał Romaszewski\corref{cor1}}
\cortext[cor1]{Corresponding author}
\ead{michal@iitis.pl}
\author[cbk]{Kamil Drejer}
\ead{kdrejer@cbk.waw.pl}
\author[iitis]{Katarzyna Kołodziej}
\ead{kkolodziej@iitis.pl}
\author[iitis]{Anna Zawadzka}
\ead{azawadzka@iitis.pl}
\author[cbk]{Stanisław Lewiński}
\ead{stlewinski@cbk.waw.pl}
\author[iitis]{Przemysław Głomb}
\ead{przemg@iitis.pl}
\author[cbk]{Marek Ruciński}
\ead{mrucinski@cbk.waw.pl}
\author[cbk]{Michal Krupiński}
\ead{mkrupinski@cbk.waw.pl}
\author[cbk]{Krzysztof Gryguc}
\ead{kgryguc@cbk.waw.pl}
\author[iitis]{Przemysław Sekuła}
\ead{psekula@iitis.pl}
\author[cbk]{Szymon Sala}
\ead{ssala@cbk.waw.pl}
 
\address[iitis]{Institute of Theoretical and Applied Informatics, Polish Academy of Sciences,\\Bałtycka 5, 44-100 Gliwice, Poland}
\address[cbk]{Space Research Centre, Polish Academy of Sciences\\Bartycka 18a, 00-716 Warszawa, Poland}

\begin{abstract}
Sentinel-2 imagery offers open access, global coverage, and frequent revisit times, making it attractive for practical building mapping at scale; however, its native $10\,\mathrm{m}$ resolution makes building vs non-building classification challenging, particularly for small or sub-pixel buildings, and performance can vary with both seasonality and the heterogeneity of built-up environments. This paper introduces a Sentinel-2 building-detection framework designed to systematically quantify these effects and to support more formalised, practice-oriented model selection. We construct a dedicated multi-temporal Sentinel-2 dataset over the Warsaw region and derive binary ground-truth masks by rasterising official Polish topographic database (BDOT10k) building footprints onto the Sentinel-2 pixel grid. Using two established convolutional segmentation backbones (U-Net and DeepLabV3+), we first perform scene-specific fine-tuning to select a robust architecture and identify the best monthly models for L1C and L2A products separately. We then conduct cross-temporal inference by applying each best monthly model to all scenes, enabling an assessment of (i) which months provide favourable training and inference conditions, (ii) how performance transfers between seasons, (iii) the impact of processing level, and (iv) how these effects differ across built-up typologies. Based on these results, we provide practical guidance for routine Sentinel-2 building classification under varying acquisition periods and settlement characteristics.
\end{abstract}

\begin{keyword}
Sentinel-2, building detection, semantic segmentation, deep learning, transfer learning
\end{keyword}
\end{frontmatter}
\section{Introduction}
With the growing volume of high-resolution satellite imagery and large-area monitoring programs, rapid and accurate extraction of built-up areas and individual buildings has become central to urban and rural management, planning, and development. Such information supports downstream analysis, including population density estimation, building-height inference \cite{dixit2021dilated}, and the identification of suitable locations for renewable energy infrastructure \cite{dixit20226+, meng2022unsupervised}. Beyond planning-oriented applications, building detection is increasingly used to monitor change and compliance by identifying structures that are illegally built, reconstructed in violation of spatial planning regulations, or abandoned \cite{reda2020detection, zou2022mapping}. For example, building extraction systems have been proposed to track housing construction on illegally occupied farmland in China \cite{li2021attention}, to map abandoned houses in shrinking cities in the United States \cite{zou2022mapping}, and to detect refugee dwellings to better estimate appropriate humanitarian aid \cite{gao2022comparing}.

Building extraction remains a complex task, shaped by several influencing factors, including spectral similarity to non-building objects, especially those within impervious surfaces; the diversity of building rooftop shapes, structures, and materials; vegetation, other rooftops, and cloud occlusions; illumination conditions \cite{rawat2024deep, dixit2021dilated, dixit20226+}. Another challenge concerns the spatial resolution of the remote sensing imagery for building extraction. Most existing datasets and research studies rely on very high-resolution (VHR) imagery ($\leq$50~cm) \cite{li2024review}, including dedicated large-scale projects such as Continental-Scale Building Detection from High Resolution Satellite Imagery \cite{sirkoContinentalScaleBuildingDetection2021} and Microsoft Global ML Footprints \cite{MicrosoftGlobalMLBuildingFootprintsWorldwide}. Although models trained on such data achieve high accuracy, their applicability is constrained by high acquisition costs, infrequent revisit cycles, limited temporal control, incomplete spatial coverage, and the scarcity of historical records \cite{rawat2024deep, li2024review, wuFlexibleFrameworkIdentifying2025, fengNationalscaleMappingBuilding2023, xuESPC_NASUnetEndtoEndSuperResolution2021}, which hinders their use in broader or long-term analyses.

In response to these limitations, there has been growing interest in leveraging medium-resolution multispectral data, particularly Sentinel-2 imagery, for building extraction. Sentinel-2 imagery is open access, provides global coverage, and offers frequent revisit times, thus supporting cost-effective analysis even in rural or remote regions \cite{dixit20226+, wuFlexibleFrameworkIdentifying2025}. At the same time, the decametre resolution of Sentinel-2 images limits the ability to accurately represent buildings, as the detected objects may be smaller than a single image pixel or occupy only small portions of adjacent pixels. Several methods address this, including sub-pixel mapping techniques that aim to determine the proportions and locations of class fractions within a pixel \cite{atkinson1997mapping, schug2022subpixel, hao2023subpixel}, as well as approaches based on the superresolution of the input data \cite{xuESPC_NASUnetEndtoEndSuperResolution2021, fengNationalscaleMappingBuilding2023, debella-giloRelativePerformanceSuperresolved2025}. Such works highlight the need for special attention to the precise identification of sub-pixel buildings. Another important yet insufficiently explored challenge in building detection is the heterogeneity of built-up areas, as only a few studies examine how detection performance varies across settlement types, regardless of their socioeconomic function. The authors of \cite{chenLargescaleIndividualBuilding2023} report significant differences in segmentation accuracy across categories such as high-rise, urban, suburban, and rural areas. Finally, despite Sentinel-2’s frequent revisit times and dense temporal coverage, the impact of seasonal variability on building detection performance remains comparatively underexplored, motivating a seasonality-focused analysis in this setting.

To better understand how heterogeneity in built-up environments affects detection performance, useful insights can be drawn from the related field of urban functional zone (UFZ) mapping, where such diversity is a central concern. UFZs are defined as spatially coherent areas, characterised by a relatively homogeneous building structure strongly associated with specific human socioeconomic activities \cite{du2024zones}. Studies show that these structural differences affect classification performance. For instance, \cite{li2024identifying} demonstrates that residential, industrial, educational, and commercial buildings are mapped more accurately than public buildings. Similarly, seasonal Sentinel-2-based features yield higher UFZ classification accuracy in residential, industrial, and green zones, and lower accuracy in commercial areas \cite{hu2025seasonal}. Research on specific contexts, such as farmland regions \cite{li2021attention}, further confirms the sensitivity of detection methods to structural variation. Overall, findings from both settlement-type and UFZ-related studies underscore that the inherent diversity of urban areas strongly conditions the performance of building extraction, suggesting the need for an explicit analysis of Sentinel-2-based detection across settlement types.

A second key factor is seasonality. Land use and land cover (LULC) classification from satellite imagery faces a fundamental challenge of temporal stability \cite{chowdhuryGISBasedMethod2023, liuSeasonalVariationLand2015}, which remains insufficiently understood in operational remote sensing workflows. Although practitioners often rely on heuristics when selecting images from a season most suitable for building detection, these choices are rarely formalised, and their effectiveness is not well quantified \cite{liuSeasonalVariationLand2015}. Recent studies highlight the benefits of incorporating seasonal variability into built-up area classification, demonstrating notable improvements across various environmental and hydrological applications, including water quality modelling, hydrological and nitrogen yield estimation, and the calibration of hydrological models \cite{myersSeasonalVariationLand2024}. In the study \cite{liDetectingHighRiseBuildings2021}, seasonal effects were taken into account when detecting high-rise buildings from Sentinel-2 imagery. However, most LULC-focused research continues to emphasise algorithmic advances \cite{ponceSatelliteImageTime, royMultitemporalLandUse2019} rather than systematically evaluating how seasonal changes affect performance across different urban morphologies. This motivates a dedicated analysis of seasonality effects in Sentinel-2-based building detection.

Urban areas with contrasting structural and environmental characteristics - high and medium density, suburban, low rise, and dispersed rural built-up areas - respond differently to seasonal changes in reflectance, illumination, shadow patterns, and vegetation coverage. Only a few studies directly examine these interactions. For example, \cite{liDetectingHighRiseBuildings2021} analyses the behaviour of the Fully Convolutional Network (FCN)-based method under various seasonal and spatial conditions, while \cite{robinsonTemporalClusterMatching2021} investigates how the colour–texture relationship between buildings and their surroundings evolves during change detection. Similarly, \cite{ettenMultiTemporalUrbanDevelopment2021} highlights that variability in image quality, atmospheric conditions, shadows, and seasonal phenology introduces additional ambiguity to building footprints. Furthermore, reviews of spatiotemporal fusion methods also note the influence of specific seasonal transitions \cite{sunDecadeDeepLearning2025}. Overall, existing work indicates that seasonality has a measurable but insufficiently characterised impact on building detection from Sentinel-2 imagery. Despite promising attempts using multitemporal data, this field still lacks a systematic assessment of how seasonal dynamics affect classification accuracy across diverse urban environments. In this study, we address this gap by introducing a building-detection framework that, by utilising a dedicated multi-temporal Sentinel-2 dataset, systematically evaluates model performance across settlement morphologies and seasonal conditions and supports the derivation of generalised, formalised guidelines for practitioners performing routine building classification with Sentinel-2 imagery.

Our contributions are as follows: 
\begin{enumerate}
    \item We propose a building-mapping framework using Sentinel-2 imagery at its native 10\,m spatial resolution.
    \item We compile a dedicated multi-temporal Sentinel-2 dataset spanning multiple settlement morphologies for evaluation.
    \item We conduct systematic cross-validation across months to identify optimal training and inference periods, quantify cross-season transfer, and determine windows of maximum classification stability, addressing a spatio-temporal generalisation gap.
    \item We provide operational guidelines for routine Sentinel-2 building mapping, including recommendations for selecting month-specific models for given acquisition periods and scene characteristics.
\end{enumerate}

\section{Materials}
\subsection{Study Area}
\label{sec:study_area}
\begin{figure}[h!]
    \centering
    \includegraphics[width=1\linewidth]{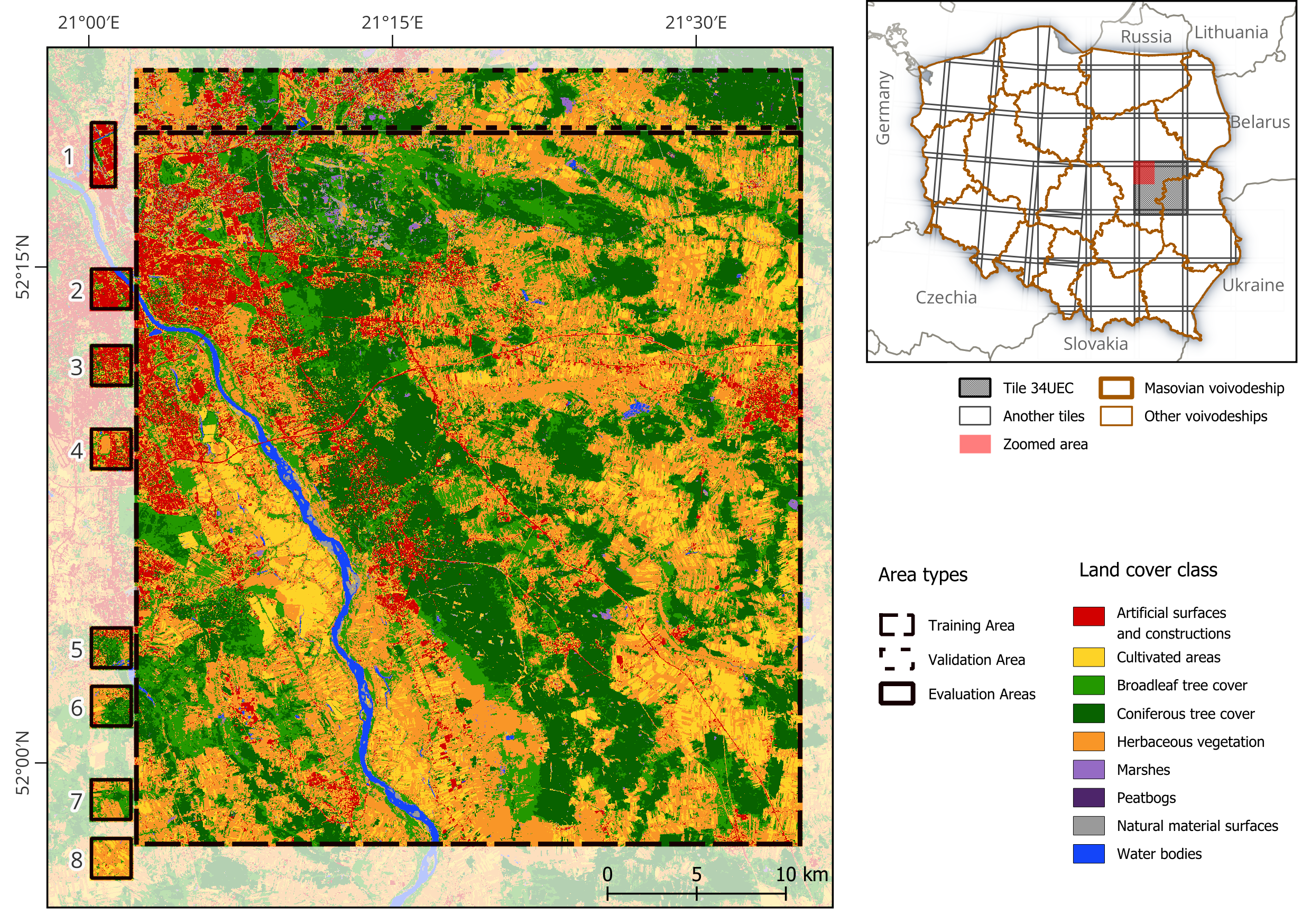}
    \caption{localisation of training, validation and evaluation (test) areas on the landcover basemap with highlighted areas representing different types of built-up environments used for evaluation: 1 -- industrial built-up area, 2 -- high density built-up area (city centre), 3 -- high density built-up area, 4 -- medium density built-up area (multi-family blocks), 5 and 6 -- suburban, low rise built-up area (single-family buildings), 7 and 8 -- dispersed rural built-up area.} 
    \label{fig:areas_map}
\end{figure}

The study area was established within the northwestern part of Sentinel-2 tile T34UEC and belongs entirely to the Mazovian Voivodeship, Poland (\Cref{fig:areas_map}).

The study area covers the central and eastern parts of the capital city of Warsaw and the adjacent areas extending southward, ensuring capture of critical local development patterns to support remote sensing analysis. From a physiogeographical perspective, it consists of plains (the Wołomin, Garwolin, and Warsaw plains) intersected by the Middle Vistula River Valley and the Warsaw Basin. Part of the area also includes the western sections of the Kałuszyn Heights. Overall, this Central Mazovian Lowland region exhibits a mix of glacial and fluvial geomorphology, diverse soils, extensive agricultural lands, forests, important river systems, including the Vistula and its tributaries, and scattered urban centres connected by significant transportation infrastructure. The study area includes protected natural environments (e.g., Natura 2000 sites) as well as sites of industrial and cultural importance. The area has a temperate climate typical of the mid-latitudes of the Northern Hemisphere, dominated by continental influences with some oceanic moderation, featuring cold winters, warm summers, and precipitation concentrated in the summer, with urban heat island effects around Warsaw \cite{richlingRegionalnaGeografiaFizyczna2021}. This diversity makes the region a suitable testbed for Sentinel-2 building detection, as it spans a dense urban fabric, suburban expansion, dispersed rural settlements, and extensive non-built backgrounds. The resulting variation in land cover, soil/vegetation context, and illumination and shadow conditions supports a robust assessment of model performance across built-up typologies and seasons.

\begin{figure}[h!]
    \centering
    \includegraphics[width=1\linewidth]{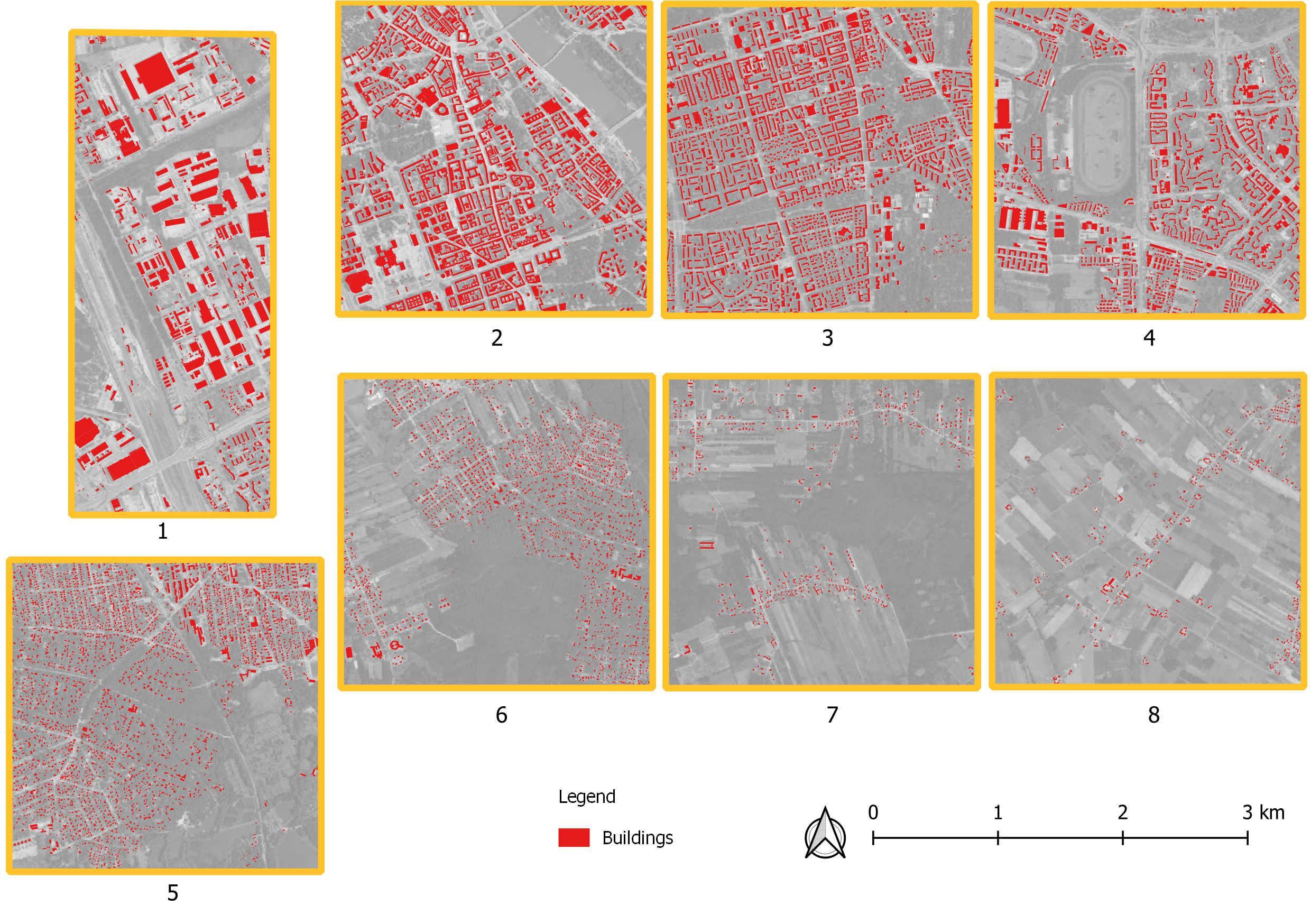}
    \caption{Selected test areas represent different types of built-up environments: 1 -- industrial built-up area, 2 -- high density built-up area (city centre), 3 -- high density built-up area, 4 -- medium density built-up area (multi-family blocks), 5 and 6 -- suburban, low rise built-up area (single-family buildings), 7 and 8 -- dispersed rural built-up area.} 
    \label{fig:areas_zoomed}
\end{figure}

Most of the study area was designated as the training area, covering approximately 1,508 km², while the validation area, located directly to its north, covers about 132 km². Eight test (evaluation) areas, covering approximately 6.25 km² each, were defined near the western boundary of the training area. They represent the basic types of built-up areas differentiated by their predominant functional type: 1 -- industrial built-up area, 2 -- high density built-up area (city centre), 3 -- high density built-up area, 4 -- medium density built-up area (multi-family blocks), 5 and 6 -- suburban, low rise built-up area (single-family buildings), 7 and 8 -- dispersed rural built-up area. The locations of the training, validation, and test areas corresponding to these numbers are shown in \Cref{fig:areas_map}. The building layouts within the test areas are visible in the enlarged sections in \Cref{fig:areas_zoomed}. 

\subsection{Data Description}

The Sentinel-2 constellation, operating in a sun-synchronous orbit with its Multi-Spectral Instrument (MSI) delivers wide-swath imagery sampling 13 spectral bands: four bands at 10 m, six bands at 20 m and three bands at 60 m spatial resolution. The dataset offers global coverage of terrestrial and coastal regions between latitudes 56° S and 82.8° N, with a nominal revisit cycle of five days at the Equator and increased frequency toward the poles. All imagery is provided under a free and open access policy, enabling unrestricted use for public, scientific, and commercial purposes \cite{Sentinel2ProductsSpecification}. Sentinel-2 satellite data, widely used for Earth observation, is distributed mainly at two processing levels: L1C and L2A. L1C products provide Top-of-Atmosphere (TOA) reflectance imagery that is radiometrically calibrated, geometrically corrected (orthorectified), and map-projected. These images serve as the primary, map-ready dataset available immediately following data  acquisition. However, they contain atmospheric effects such as haze, aerosols, and cloud interference without any correction. L2A products advance beyond L1C by applying atmospheric correction (using the Sen2Cor algorithm) to deliver Bottom-of-Atmosphere (BOA) reflectance, representing surface reflectance values. The L2A products have 12 spectral bands - cirrus band 10 is omitted, as it does not contain surface information  \cite{Sentinel2ProductsSpecification}\cite{gintingComparisonTopBottom2024}. 

The models used in the experiment for both processing levels use all 13 spectral bands (B01-B12, including B8A) of the L1C data. For L2A data, an empty band was inserted to preserve the band order consistent with the pretrained model ~\cite{wangDecouplingCommonUnique2024a}. All spectral bands were resampled to a spatial resolution of 10m using the nearest-neighbour interpolation method. Digital Numbers (DN) were converted to Surface Reflectance (SR) values following the Sentinel-2 documentation \cite{Sentinel2ProductsSpecification}, using the transformation $SR = (DN - 1000)/10000,$ and clipped to the $[0, 1]$ range.

The experimental dataset consists of 130 Sentinel-2 images corresponding to 65 acquisition dates, available at both L1C and L2A processing levels, and selected under low-cloud conditions in the analysed region. Scenes were selected from a broad temporal range, 2019-2025, to ensure sufficient material for robust analysis, as the exclusion of cloud-contaminated imagery substantially reduces the set of available scenes. Characteristics of the data used in the experiment are presented in \Cref{tab:data-characteristics}. The summary of the number of selected scenes for individual months and years is presented in \Cref{tab:scenes_matrix}. 

\begin{table}[h!]
\centering
\begin{tabular}{|c|>{\raggedright}p{4cm}|>{\raggedright\arraybackslash}p{8cm}|}
\hline
\textbf{lp} & \textbf{parameter} & \textbf{value} \\ \hline
1 & time period & 04.2019 -- 07.2025 \\ \hline
2 & processing levels & L1C, L2A \\ \hline
3 & tiles & T34UEC \\ \hline
4 & bands & L1C: [1, 2, 3, 4, 5, 6, 7, 8, 8A, 9, 10, 11, 12] \newline
            L2A: [1, 2, 3, 4, 5, 6, 7, 8, 8A, 9, 11, 12] \\ \hline
5 & cloud cover & $<$20\% and additional manual selection to minimize the cloud coverages in AOIs \\ \hline
\end{tabular}
\caption{Input Sentinel-2 data parameters}
\label{tab:data-characteristics}
\end{table}

\begin{table}[h!]
\centering
\small
\setlength{\tabcolsep}{4pt}
\renewcommand{\arraystretch}{1.1}
\begin{tabular}{l|cccccccccccc|c}
\hline
\textbf{Year} & Jan & Feb & Mar & Apr & May & Jun & Jul & Aug & Sep & Oct & Nov & Dec & \textbf{Total} \\
\hline
2019 & -- & 1 & -- & 2 & -- & 2 & 1 & 2 & 1 & 2 & -- & 2 & 13 \\
2020 & -- & -- & 2 & 2 & 1 & -- & -- & 1 & 1 & -- & 1 & -- & 8 \\
2021 & -- & -- & -- & 1 & 1 & -- & -- & 1 & -- & 1 & 2 & 1 & 7 \\
2022 & -- & 1 & 2 & -- & 2 & 1 & 2 & 1 & 1 & -- & 1 & -- & 11 \\
2023 & -- & -- & -- & -- & -- & -- & -- & 1 & 4 & -- & -- & 1 & 6 \\
2024 & 2 & -- & 1 & -- & 2 & 2 & -- & -- & 1 & 2 & 1 & -- & 11 \\
2025 & -- & 1 & 4 & 3 & -- & -- & 1 & -- & -- & -- & -- & -- & 9 \\
\hline
\textbf{Total} & 2 & 3 & 9 & 8 & 6 & 5 & 4 & 6 & 8 & 5 & 5 & 4 & 65 \\
\hline
\end{tabular}
\caption{Matrix of Sentinel-2 acquisition dates used in the analysis. Values indicate the number of scenes per month and year selected according to the specified criteria (see text). The last column and last row show yearly and monthly totals.}
\label{tab:scenes_matrix}
\end{table}

Ground-truth data was derived from BDOT10k \footnote{https://www.geoportal.gov.pl/en/data/topographic-objects-database-bdot10k}, Poland’s official topographic vector database, developed and maintained under strict national regulations and characterised by high levels of consistency, reliability and attribute richness. Most features are sourced from the national Land and Building Register (EGiB \footnote{https://www.geoportal.gov.pl/en/data/land-and-building-register-egib}), ensuring high geodetic accuracy. It is also based on additional sources, including orthophotomaps and on-site inspections. The BDOT10k, particularly in rural areas, exhibits higher attribute and areal completeness than in OpenStreetMap (OSM) – a collaborative project that creates a free, editable map of the world, built by volunteers \cite{borkowskaOpenStreetMapBuildingData2023}. In BDOT10k, building geometries are captured with at least 1.5~m accuracy and coordinates precise to 0.01~m. The ‘building’ class includes all residential and non-residential structures greater than 40~m², as well as smaller non-residential buildings of significance or which are a part of a row of buildings. Temporary, small, or garden structures are excluded \cite{poland_law_2021_item1412}. The database is updated progressively every 3–5 years. Based on information provided by the Polish National Geoportal, the data in the analysis area, depending on location, come from the years 2022-2024 \cite{Geoportalgovpl}.

Applied classification algorithms (\Cref{sec:model_characteristics},  \Cref{sec:hyperparameters_setup}) process input data with 10 m pixels and generate classification maps with 3.5x higher spatial resolution (2.857 m).

The ground-truth data was generated by rasterising the vector BDOT10k database into a binary mask:
\begin{itemize}
    \item  with 10 m pixel, aligned with the Sentinel-2 pixel grid for training,
    \item with 2.857 m pixel, aligned with inference results for testing.
\end{itemize}

\begin{figure}[h!]
    \centering;
    \includegraphics[width=1\linewidth]{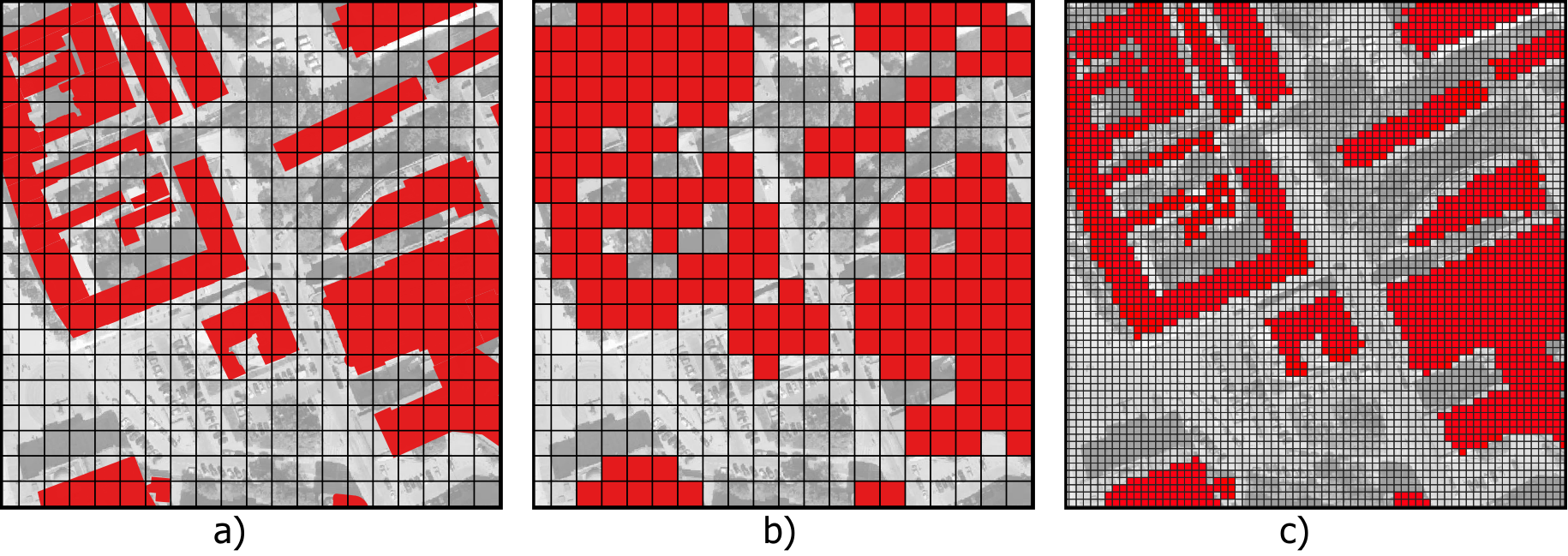}
    \caption{Example of ground truth data. a) vector database derived from BDOT10k database, b) mask rasterized in 10m resolution, harmonised with the Sentinel-2 scenes c) mask rasterized in 2.875m resolution, harmonized with the results grid. Built-up areas are shown in red; the grid depicts an input 10m or a result 2.857m pixel layout.} 
    \label{fig:1x3_maski}
\end{figure}

Grid cell occupancy thresholds ranging from 10\% to 100\% (in 10\% increments) were investigated to represent vector footprints in 10 m rasterisation. A 30\% threshold was adopted because it provided the most faithful representation of building shapes while minimising distortion of footprint geometry. Visualisation of the example ground truth and the resulting generalisation is shown in \Cref{fig:1x3_maski}.

To ensure a comprehensive evaluation of the results, the test dataset was additionally compared with historical BDOT10k datasets and archival aerial imagery. In places, where changes in building footprints occurred between 2019 and 2025 – such as new construction, demolition or rebuilding -- the relevant area was excluded from the evaluations. The test dataset was supplemented with additional structures not part of the BDOT10k building dataset, identified using other BDOT10k layers and aerial imagery, including greenhouses, parking area shelters, glazed road sound barriers, and clearly visible auxiliary elements such as patios. This approach minimises the effects of temporal discrepancies between ground-truth and satellite data during quality assessment and better aligns the evaluation with the actual state of the observed features.

\section{Methods}
\label{sec:methods}
Sentinel-2 building detection is affected by seasonal variability (vegetation, illumination, shadows) and by the heterogeneity of built-up environments. The experiment described below quantifies these effects by evaluating model performance and stability across months and settlement morphologies, and by measuring cross-seasonal transferability.

\begin{figure}
    \centering
    \includegraphics[width=1\linewidth]{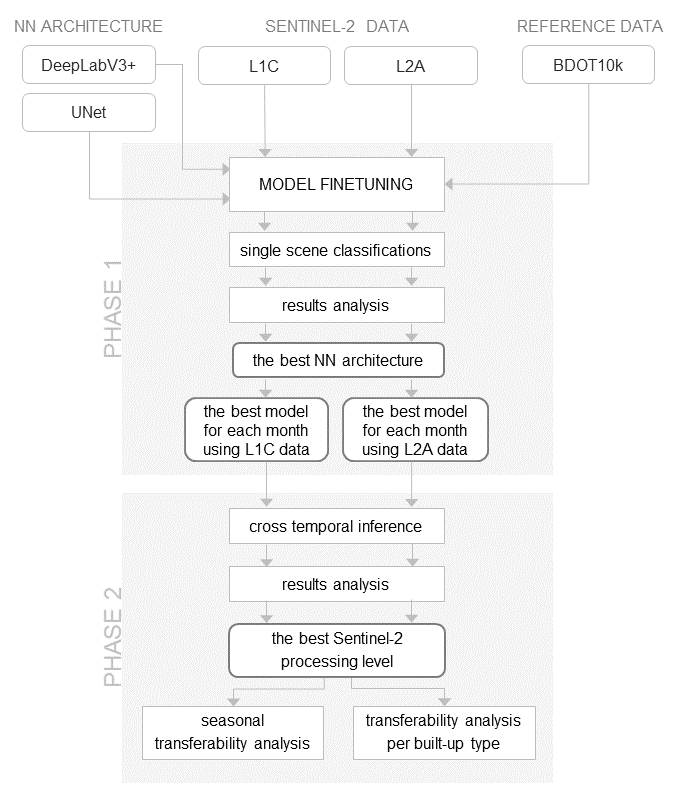}
    \caption{General workflow of experimental pipeline. Phase~1 performs scene-specific fine-tuning of UNet and DeepLabV3+ to select the best-performing architecture and the best monthly models separately for Sentinel-2 L1C and L2A products. Phase~2 applies cross-temporal inference by classifying all scenes with each best monthly model to evaluate processing-level effects and seasonal transferability across built-up area types.} 
    \label{fig:scheme}
\end{figure}

The experiment was performed in two main phases. Phase 1 focused on scene-specific fine-tuning, in which each Sentinel-2 scene was used to train a dedicated model using two convolutional architectures: UNet~\cite{ronnebergerUNetConvolutionalNetworks2015} and DeepLabV3+ \cite{chenEncoderDecoderAtrousSeparable2018}. 
From the resulting set of fine-tuned models, the best-performing architecture was identified. Next, for this selected architecture, the best model per month was selected separately for L1C and L2A products, using the validation data subset to avoid data leakage in the next Phase. In Phase 2, cross-temporal inference was performed by classifying test areas across all scenes using each of the best monthly models for both S2 product types. These results helped determine the recommended level of processing (L1C or L2A) for Sentinel-2 imagery for cross-temporal applications. In the final step of the analysis, the seasonal transferability of the classification models in different types of built-up areas was comprehensively examined. The general experiment workflow is presented in \Cref{fig:scheme}.
 
\subsection{Model characteristics}
\label{sec:model_characteristics}
We used an encoder-decoder architecture for binary semantic segmentation, with a ResNet-50 encoder and UNet/DeepLabV3+ decoder. U-Net and DeepLabV3+ were selected as reliable, widely adopted baselines to keep the focus on spatio-temporal generalisation rather than architectural novelty. U-Net emphasizes spatial detail through skip connections, which is important for buildings at $10\,\mathrm{m}$ resolution, whereas DeepLabV3+ explicitly captures multi-scale contextual information via atrous convolutions and refines boundaries in its decoder. This complementary pairing helps ensure that conclusions about seasonality and settlement-type effects are not driven by idiosyncrasies of a single model. 

The encoder was initialised with weights pretrained on Sentinel-2 data (L1C and L2A) using the DeCUR approach~\cite{wangDecouplingCommonUnique2024a}. During fine-tuning, the encoder parameters were frozen, and only the decoder and segmentation head were updated to reduce the risk of overfitting. We used cross-entropy loss, the Adam optimiser, and a cosine annealing learning rate scheduler.

\subsection{Hyperparameters and computational setup}
\label{sec:hyperparameters_setup}
Input patches of 64$\times$64 pixels were resampled to $224 \times 224$ pixels resulting in output value of approximately 2.86\,m. The training area was subdivided into patches using random sampling. The network was trained for 600 epochs with a batch size of 32. The initial learning rate was set to $10^{-3}$, with a minimum of $10^{-5}$ defined in the scheduler. The architecture demonstrated robustness with respect to moderate hyperparameter variations. The implementation was carried out in Python~3.12.1 using the PyTorch~2.2.0 and TorchGeo\footnote{\url{https://www.osgeo.org/projects/torchgeo}}~0.6.2 libraries.

During fine-tuning, the encoder was frozen, and only the respective decoders were trained, which reduced compute. This means we are training approximately. $28\%$ of $\sim32$M total parameters for UNet and approx. $12\%$M of $\sim27$ total parameters for DeepLabV3+. For inference throughput, UNet reaches $\sim830$ and DeepLabV3+ $\sim890$ samples/second with similar peak VRAM $\sim950$MB on a computer with Intel i9-10980XE CPU and twin GeForce RTX 3090 GPUs.

\subsection{Classification}
In Phase 1, each Sentinel-2 image was classified with a dedicated fine-tuned model, which resulted in 2 model architectures $\times$\ 2 Sentinel-2 processing levels (L1C and L2A) $\times$\ 65 dates $\times$\ 8 test areas yielding a total of 2080 classifications. This phase was intended to identify the better-performing model architecture and, for each processing level, to select the best monthly model for each month from January to December. 

In Phase 2, each Sentinel-2 image was classified with the best monthly model from Phase 1. It results in 1 model architecture $\times$\ 12 best monthly models $\times$\ 2 Sentinel-2 processing levels (L1C and L2A) $\times$\ 65 dates $\times$\ 8 test areas producing 12480 classifications. Accuracy analyses of these maps were conducted to provide insight into the models’ cross-temporal transferability.

\subsection{Evaluation}
Ground truth with a spatial resolution of classification results (2.857 m) was used to calculate the performance metrics for each test area:
\begin{itemize}
    \item \textbf{IoU (Intersection over Union)} - measures the overlap between predicted and reference building pixels as the ratio of their intersection to their union:
    \[
    \mathrm{IoU} = \frac{TP}{TP + FP + FN}
    \]
    
    \item \textbf{F1-score} - the harmonic mean of precision and recall, capturing the balance between false positives and false negatives:
    \[
    \mathrm{F1} = 2 \cdot \frac{\mathrm{Precision} \cdot \mathrm{Recall}}{\mathrm{Precision} + \mathrm{Recall}} = \frac{2TP}{2TP + FP + FN}
    \]
    
    \item \textbf{BA (Balanced Accuracy)} - accounts for class imbalance by averaging the recall obtained on each class:
    \[
    \mathrm{Balanced\ Accuracy} = \frac{1}{2} \left( \frac{TP}{TP + FN} + \frac{TN}{TN + FP} \right)
    \]
    where $TN$ denotes true negatives, $TP$ is true positives, $FP$ is false positives, and $FN$ is false negatives, $Precision$ is $TP/(TP+FP)$, and $Recall$ is $TP/(TP+FN)$.
\end{itemize}

All cloud-covered areas were omitted from the evaluation of each classification. The cloud mask was generated using the \texttt{s2cloudless} \cite{SentinelhubSentinel2clouddetector2026} library with the following parameters: 
\texttt{threshold=0.7}, 
\texttt{average\_over=4}, 
\texttt{dilation\_size=2}, 
\texttt{all\_bands=True}. 
Classification results with cloud cover exceeding 60\% were excluded from the evaluation to avoid potential distortion of the performance metrics.

\subsection{Methodology of separating detection from localisation error}

To separate interior misclassification from boundary disagreement expected from the resolution gap between Sentinel-2 and the building size, we additionally decomposed the reference mask using morphological operations.

We analyse this effect by applying one-pixel morphological erosion \(\varepsilon(\cdot)\) and dilation \(\delta(\cdot)\) to the reference building mask \(G\), which yields four complementary regions: the eroded building core \(G_c=\varepsilon(G)\), the inner boundary band \(G_{bi}=G\setminus\varepsilon(G)\), the outer boundary band \(G_{bo}=\delta(G)\setminus G\), and the far background \(G_f=\Omega\setminus\delta(G)\), where \(\Omega\) is the image domain. Using the binary prediction mask \(P\), we then decompose the confusion matrix into interior and boundary components: \(TP_i=|P\cap G_c|\), \(FN_i=|(\Omega\setminus P)\cap G_c|\), \(TP_b=|P\cap G_{bi}|\), \(FN_b=|(\Omega\setminus P)\cap G_{bi}|\), \(TN_b=|(\Omega\setminus P)\cap G_{bo}|\), \(FP_b=|P\cap G_{bo}|\), \(TN_i=|(\Omega\setminus P)\cap G_f|\), and \(FP_i=|P\cap G_f|\). This simple construction isolates agreement in the stable interior of building footprints from disagreement in the immediate \(\pm 1\) px boundary neighbourhood and from false positives in unambiguous background regions.

To keep this analysis aligned with the main evaluation protocol, we use the two principal metrics from elsewhere in the paper, Balanced Accuracy and F1, and compute them for both the standard full-image confusion matrix and the morphology-restricted regions defined above. In particular, our key summary metric is the guardrail balanced accuracy, defined analogously to the standard case but using only the stable building core and far background, i.e. \(BA_{\mathrm{guard}}=\tfrac{1}{2}\left(\tfrac{TP_i}{TP_i+FN_i}+\tfrac{TN_i}{TN_i+FP_i}\right)\). This measure answers how well the model performs once the immediate boundary uncertainty is excluded. To characterise localisation at the building contour itself, we additionally use boundary recall, \(R_{\mathrm{bnd}}=\tfrac{TP_b}{TP_b+FN_b}\), which measures agreement in the inner one-pixel boundary band, and boundary specificity, \(S_{\mathrm{bnd}}=\tfrac{TN_b}{TN_b+FP_b}\), which measures suppression of false positives in the outer one-pixel band. Finally, to quantify how much of the total error content is concentrated near building edges, we use the boundary error share, defined as \((FN_b+FP_b)/(FN+FP)\). Together, these metrics provide a compact description of three complementary aspects: overall performance under the standard protocol, performance in the unambiguous interior-versus-background setting, and localisation quality in the immediate boundary zone.

\section{Results} 
Results were obtained for Sentinel-2 imagery from 2019–2025. The test area consists of eight subsets corresponding to the following types of built-up areas: industrial, dense urban, medium-density residential, suburban, and dispersed rural. The accuracy of the obtained built-up masks is reported using three metrics: IoU, F1-score, and BA, enabling assessment under class-imbalance conditions. 

Because Sentinel-2 building detection is affected by several interacting factors, the Results section is organised to separate these effects. The investigation involves the model architecture, the Sentinel-2 processing level, the acquisition season, the structure of the built-up area, and the spatial mismatch between imagery and building-footprint reference data. We evaluate:

\begin{enumerate}
    \item \textbf{Model configuration and data preprocessing}: Section~\ref{sec:results_phase_1_model} compares U-Net and DeepLabV3+ under the same experimental protocol -- this is described as Phase-1 in Methodology Section~\ref{sec:methods}. Section~\ref{sec:results_phase_1_accuracy} then selects the best monthly models separately for L1C (top-of-atmosphere reflectance) and L2A (atmospherically corrected surface reflectance) products.  This step also indicates whether atmospheric correction leads to a measurable improvement in this task.

    \item \textbf{Temporal behaviour}: Section~\ref{sec:results_phase_2} applies the selected monthly models to images acquired across the full 2019--2025 period. Section~\ref{sec:crosstemp_monthly} reports performance when the model and classified image correspond to the same calendar month, while Section~\ref{sec:crosstemp_seasonal} examines seasonal transferability, i.e. how well a model selected for one month performs when applied to imagery from other months.

    \item \textbf{Impact of built-up area on temporal effects}:  Section~\ref{sec:crosstemp_settlement} compares performance across dense urban, suburban, industrial, and dispersed rural test areas. Section~\ref{sec:crosstemp_quantitative} complements the pixel-wise metrics by evaluating total built-up area estimation.

    \item \textbf{Boundary-localisation and spatial-transfer errors}: Section~\ref{sec:loc_error} separates errors occurring in stable building interiors and background areas from errors located near building boundaries, where disagreement is possible because Sentinel-2 pixels are much coarser than the reference footprints. Section~\ref{spatial_transferability} then provides an external spatial-transfer check by applying the selected approach outside the main test region.
\end{enumerate}

Our investigation answers four practical questions: which architecture and processing level to use, which acquisition periods are most reliable, how strongly performance depends on the settlement structure, and how much of the observed error represents true detection failure rather than boundary uncertainty or spatial transfer limitations.

\subsection{Selection of model architecture}
\label{sec:results_phase_1_model}
\begin{figure}[H]
    \centering
    \includegraphics[width=1\linewidth]{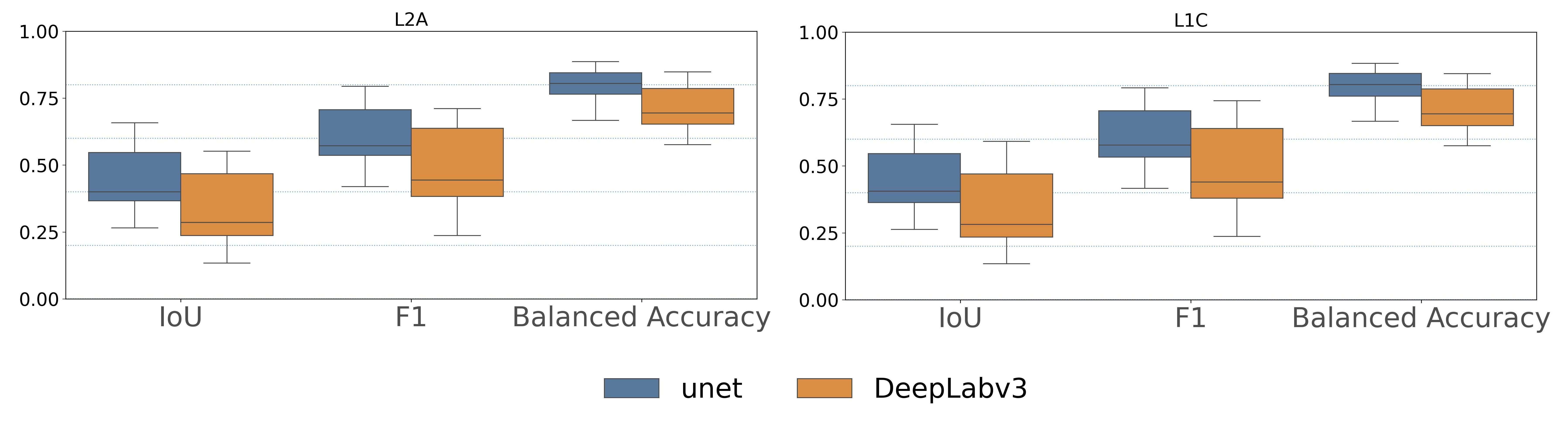}\hfill
    \caption{Distribution of performance metrics (IoU, F1-score, and Balanced Accuracy) for UNet and DeepLabV3+ in Phase 1 for single scene classifications. Left - for L2A, right - L1C products.}
    \label{fig:fig1_unet_vs_deeplabv3v2}
\end{figure}

Within Phase 1, we evaluated UNet and DeepLabV3+ using the same scene-fine-tuning and inference procedure. \Cref{fig:fig1_unet_vs_deeplabv3v2} presents, for both models, the distribution of performance metrics across all satellite data acquisition dates, processing levels, and built-up types.

All three accuracy metrics showed improved performance when the UNet architecture was used. Median IoU values for UNet were 43\%, F1 31\%, and BA 15\% higher than those of DeepLabV3+ for L1C products. Moreover, UNet exhibits narrower interquartile ranges (IQRs) for all three metrics. Mean performance results did not indicate significant differences between L1C and L2A products.

The superior performance of U-Net in this experiment likely reflects a better match between its architecture and the demands of urban building classification. In this task, the main errors usually involve building borders, small or narrow buildings, confusion among adjacent roofs, courtyards, and roads, and fragmented predictions. Through its encoder-decoder design with skip connections, U-Net preserves fine spatial detail, supports sharper boundaries, and enables better recovery of small structures. This is consistent with the higher IoU and F1 scores, since both metrics are sensitive to overlap errors, boundary imprecision, and local omission errors.

DeepLabV3+, in contrast, emphasises context aggregation through atrous convolutions and atrous spatial pyramid pooling (ASPP). While this is advantageous when broader semantic context is important, it can also produce smoother, more spatially diluted predictions. In building mapping, this may lead to less precise boundaries, merging of nearby objects, and under- or over-segmentation. The higher balanced accuracy of U-Net supports this interpretation, suggesting that it may both miss fewer building pixels and produce fewer false building detections in background areas. This is plausible in urban scenes, where the target class is more strongly defined by geometry, edges, and local shape cues than by global image structure.

The narrower IQR obtained by U-Net is also important, as it indicates not only better average performance but also greater stability across scenes. A likely reason is that U-Net provides a more direct and robust path from input detail to output mask, which makes it easier to optimise reliably. DeepLabV3+ is generally more difficult to use well because its performance depends more strongly on hyperparameter choices such as output stride, dilation rates, backbone, batch normalisation with small batch sizes, and the relation between tile size and effective receptive field.

Given its consistent superiority across metrics and seasons, UNet is selected as the reference architecture for all subsequent analyses in Phase 2, along with both Sentinel-2 product types, L1C and L2A.  

\subsection{Classification performance for L1C and L2A processing levels}
\label{sec:results_phase_1_accuracy}
Before evaluating temporal transferability, we first estimate the performance achieved during model selection. For each calendar month and each Sentinel-2 processing level, all available scenes from that month were used to train candidate models on the training subset and evaluate them on the validation subset. The model with the best validation performance was then selected as the representative monthly model for the subsequent cross-temporal analysis. \Cref{tab:l1c_l2a_months} reports these selected models, their acquisition dates, and validation metrics. These values represent selection-stage, optimistic performance estimates: they indicate the expected performance when the model is selected for a given month and product type, before it is applied to independent scenes.

For the L1C product, the highest IoU and F1-score were obtained in December (0.420 and 0.574, respectively), while the highest BA was recorded in September (0.804). In contrast, the lowest IoU occurred in April (0.391), and the lowest F1-score was observed in May (0.537). The lowest BA was recorded in June (0.786).

For the L2A product, the highest IoU was achieved in February (0.426), whereas the highest F1-score was observed in both February and October (0.580). However, August and September achieved very similar F1-scores (0.579). The highest BA was obtained in September (0.806). All three metrics - IoU, F1 and BA recorded the lowest values in January (0.400, 0.555 and 0.787, respectively).

The accuracy levels derived for both product types (L1C and L2A) are highly consistent. Any differences are minimal, typically occurring in the second or, more often, the third decimal place. The count of “wins” suggests that Sentinel‑2 imagery processed with atmospheric correction (L2A) produced marginally better outcomes.

\begin{table}[ht]
\centering
\small
\caption{Best monthly U-Net models selected in Phase~1 for Sentinel-2 L1C and L2A products. The table reports the acquisition date of each selected scene and the corresponding validation metrics (IoU, F1-score, and BA). These models are used for cross-temporal inference in Phase~2.}
\label{tab:l1c_l2a_months}
\begin{tabular}{lcccccccc}
\hline
& \multicolumn{4}{c}{\textbf{L1C}} & \multicolumn{4}{c}{\textbf{L2A}} \\
\cline{2-5}\cline{6-9}
\textbf{Month} 
& \textbf{Date} & \textbf{IoU} & \textbf{F1} & \textbf{BA}
& \textbf{Date} & \textbf{IoU} & \textbf{F1} & \textbf{BA} \\
\hline
January   & 2024-01-29 & 0.399 & 0.554 & 0.789 & 2024-01-29 & 0.400 & 0.555 & 0.787 \\
February  & 2022-02-28 & 0.401 & 0.555 & 0.792 & 2022-02-28 & \textbf{0.426} & \textbf{0.580} & 0.804 \\
March     & 2022-03-25 & 0.410 & 0.563 & 0.780 & 2024-03-09 & 0.415 & 0.570 & 0.779 \\
April     & 2025-04-03 & 0.391 & 0.544 & 0.788 & 2020-04-19 & 0.419 & 0.573 & 0.794 \\
May       & 2021-05-09 & 0.392 & 0.537 & 0.799 & 2021-05-09 & 0.421 & 0.575 & 0.801 \\
June      & 2024-06-17 & 0.398 & 0.552 & 0.786 & 2024-06-17 & 0.416 & 0.570 & 0.800 \\
July      & 2025-07-02 & 0.406 & 0.561 & 0.795 & 2025-07-02 & 0.399 & 0.554 & 0.790 \\
August    & 2023-08-12 & 0.405 & 0.558 & 0.795 & 2023-08-12 & 0.425 & 0.579 & 0.804 \\
September & 2023-09-06 & 0.412 & 0.565 & \textbf{0.804} & 2023-09-06 & 0.425 & 0.579 & \textbf{0.806} \\
October   & 2021-10-31 & 0.410 & 0.565 & 0.793 & 2021-10-31 & 0.425 & \textbf{0.580} & 0.800 \\
November  & 2021-11-10 & 0.405 & 0.561 & 0.790 & 2020-11-05 & 0.414 & 0.570 & 0.792 \\
December  & 2021-12-25 & \textbf{0.420} & \textbf{0.574} & 0.787 & 2021-12-25 & 0.422 & 0.576 & 0.789 \\
\hline
\end{tabular}
\end{table}

\subsection{Cross-temporal inference  (Phase 2)}
\label{sec:results_phase_2}
Within Phase~2, the best-performing monthly models selected in Phase~1 were applied to Sentinel-2 scenes acquired across the full study period to assess temporal transferability. Unless stated otherwise, the reported metrics are computed on the independent test areas and reported as medians across the relevant scenes and regions for each analysis. The results are analysed by acquisition month (\Cref{sec:crosstemp_monthly}), by seasonal transfer between model and image months (\Cref{sec:crosstemp_seasonal}), and across built-up types (\Cref{sec:crosstemp_settlement}). Quantitative estimates for individual test areas are presented in \Cref{sec:crosstemp_quantitative}.

\subsubsection{Monthly models performance}
\label{sec:crosstemp_monthly}
\begin{figure}[H]
    \centering
    \includegraphics[width=1\linewidth]{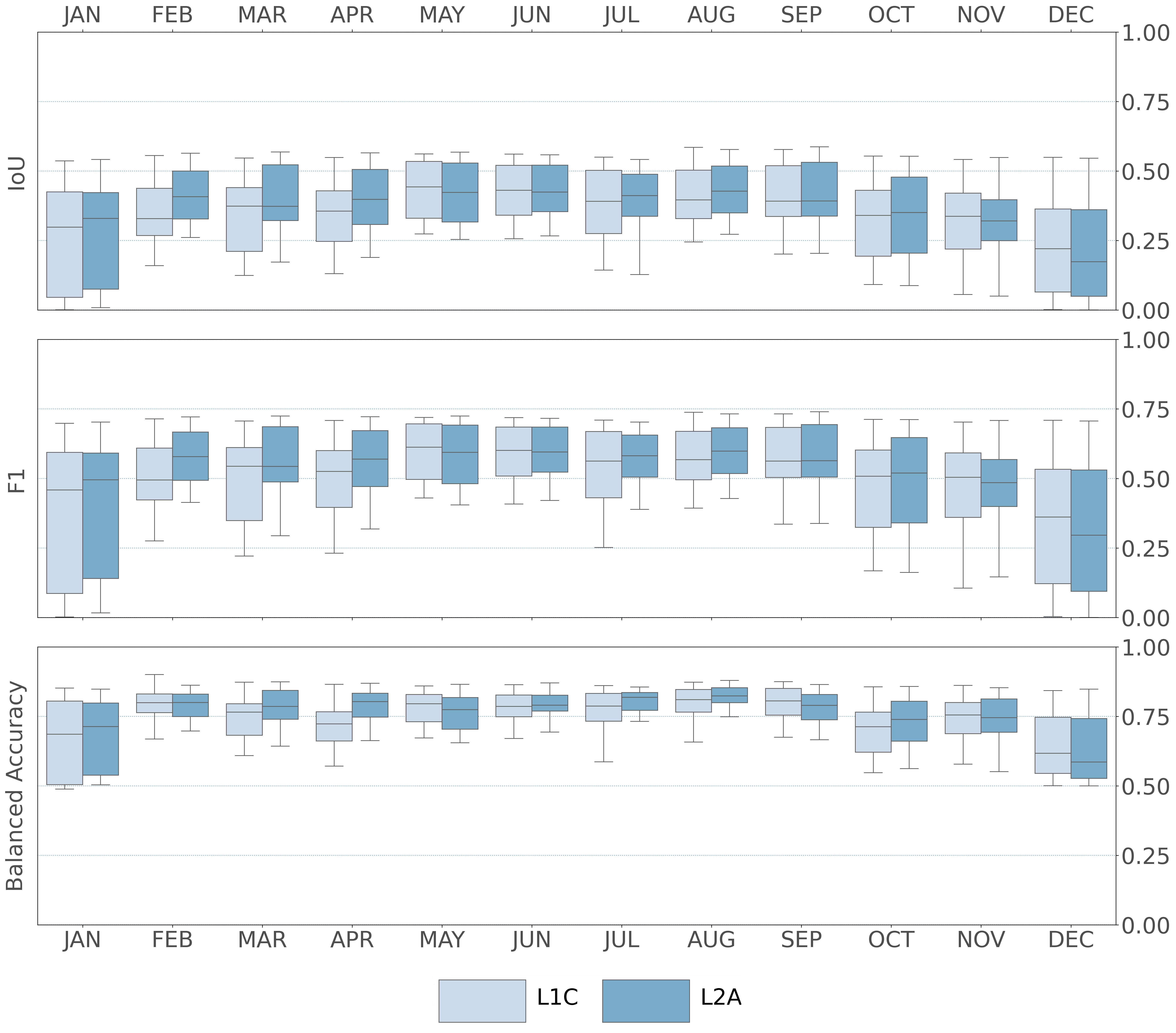}
    \caption{Monthly distributions of Phase~2 classification performance for Sentinel-2 L1C and L2A products. For each calendar month, test scenes acquired in that month were classified using the corresponding best monthly model selected in Phase~1 for the same processing level. Boxplots show IoU, F1-score, and balanced accuracy (BA) across the evaluated test scenes and areas.}
    
    \label{fig:fig2one}
\end{figure}

The performance of the best monthly models has been evaluated using test-region images acquired in the corresponding calendar month across 2019–2025. This means that all images from, e.g. January are classified with the best model for January chosen in the previous step. Therefore, the performance of the models was evaluated for the relevant time of year. The accuracy results gathered within one boxplot per month, separately by L1C and L2A, are presented in \Cref{fig:fig2one}. Numerical values for all metrics are gathered in \Cref{tab:l1c_l2a_months_met}.

Comparison with the validation results from Phase~1 (\Cref{tab:l1c_l2a_months}) shows that the monthly models do not perform uniformly across independent test scenes. For several late-spring and summer months, the Phase~2 performance remains close to, or even slightly exceeds, the validation-stage values. This is most visible for L1C in May and June, where IoU and F1 increase relative to the selected-scene validation results, and for L2A in July and August, where the metrics remain very similar or improve slightly. In contrast, the largest discrepancies occur in winter, especially in December. For L2A, IoU decreases from 0.422 to 0.174 and F1 from 0.576 to 0.296; for L1C, IoU decreases from 0.420 to 0.221 and F1 from 0.574 to 0.362. January also shows a clear reduction, although smaller than December. 

\begin{table}[H]
\centering
\small
\caption{Monthly median IoU, F1, and BA for Sentinel-2 L1C and L2A products in Phase 2, based on scenes acquired in a given month and classified with the models for that month and processing level selected in the previous step. Bold indicates the highest value within each column.}
\label{tab:l1c_l2a_months_met}
\begin{tabular}{lcccccc}
\hline
& \multicolumn{3}{c}{\textbf{L1C}} & \multicolumn{3}{c}{\textbf{L2A}} \\
\cline{2-4}\cline{5-7}
\textbf{Month}
& \textbf{IoU} & \textbf{F1} & \textbf{BA}
& \textbf{IoU} & \textbf{F1} & \textbf{BA} \\
\hline
January   & 0.298 & 0.459 & 0.686 & 0.329 & 0.495 & 0.713 \\
February  & 0.329 & 0.495 & 0.799 & 0.408 & 0.579 & 0.800 \\
March     & 0.374 & 0.544 & 0.765 & 0.373 & 0.543 & 0.786 \\
April     & 0.356 & 0.525 & 0.723 & 0.398 & 0.569 & 0.803 \\
May       & \textbf{0.443} & \textbf{0.613} & 0.796 & 0.423 & 0.594 & 0.774 \\
June      & 0.431 & 0.601 & 0.786 & 0.425 & 0.595 & 0.790 \\
July      & 0.391 & 0.562 & 0.787 & 0.411 & 0.582 & 0.819 \\
August    & 0.396 & 0.567 & \textbf{0.810} & \textbf{0.427} & \textbf{0.598} & \textbf{0.824} \\
September & 0.392 & 0.563 & 0.806 & 0.393 & 0.564 & 0.790 \\
October   & 0.341 & 0.508 & 0.713 & 0.351 & 0.519 & 0.739 \\
November  & 0.337 & 0.504 & 0.755 & 0.321 & 0.485 & 0.745 \\
December  & 0.221 & 0.362 & 0.617 & 0.174 & 0.296 & 0.586 \\
\hline
\end{tabular}
\end{table}

Overall, classification accuracy exhibits a clear seasonal pattern. The highest median values are observed from May to September, with peak performance in August and September, where both IoU and F1 reach their maximum levels, and BA exceeds 0.80 (for L2A in August). These months also show reduced standard deviations, particularly for L2A.

In contrast, the weakest and most variable results occur during the winter season (December–January). December records the lowest mean IoU and F1 for both processing levels with large standard deviations (e.g., IoU ± 0.171–0.182), indicating substantial inter-scene variability, including snow cover in half of the classified scenes.

Periods of seasonal transition (early spring and autumn) also demonstrate reduced effectiveness compared to peak summer months. In particular, October shows a noticeable drop in performance and increased variability in relation to summer.

Results indicate that classifications based on L2A imagery generally outperform those based on L1C, particularly from late winter to spring (February–April) and in October. During these months, L2A consistently yields higher IoU, F1-score, and BA values. Moreover, summer months' results show both higher means and lower variability, indicating improved predictability.  

\subsubsection{Seasonal transferability analysis}
\label{sec:crosstemp_seasonal}
\begin{figure}[H]
    \centering
    \includegraphics[width=1\linewidth]{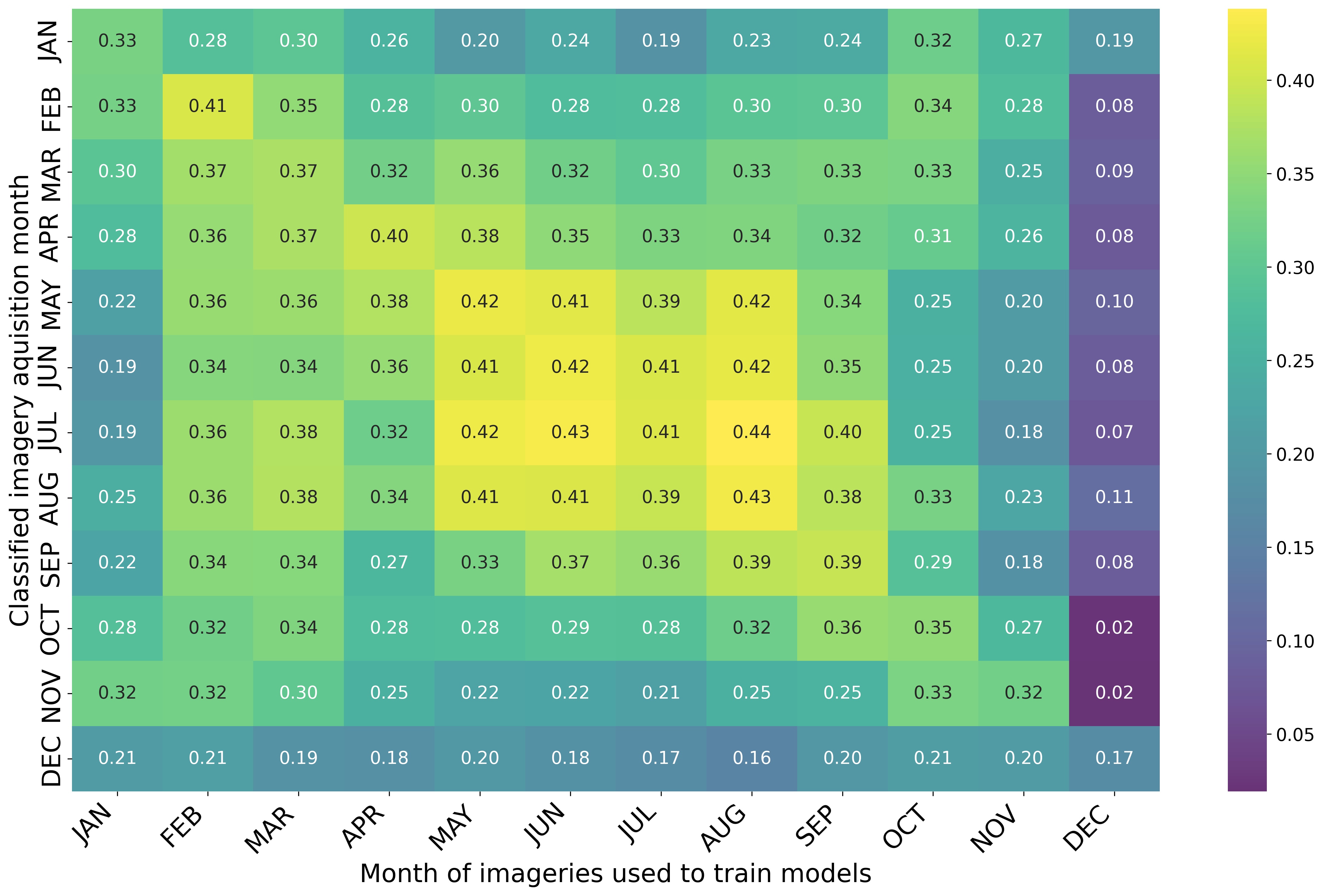}
    \caption{Seasonal transferability of the selected monthly models for Sentinel-2 L2A products. Values show the median IoU for each combination of model month (columns) and classified-scene acquisition month (rows). Diagonal cells correspond to month-matched inference, while off-diagonal cells indicate transfer to other months.}
    \label{fig:fig_IoU_transferability_heatmap_phase2}
\end{figure}

Seasonal transferability was assessed by applying each selected monthly model to scenes acquired in all other months. This analysis tests whether a model selected for one part of the year remains reliable when acquisition conditions change, which is important when choosing an appropriate model or training scene for operational mapping. The results are shown for L2A products in \Cref{fig:fig_IoU_transferability_heatmap_phase2}, because L2A generally provided slightly better and more stable performance in the preceding analyses. The matrix reports median IoU values, with columns corresponding to the selected model's month and rows to the acquisition month of the classified scenes. The diagonal represents month-matched inference, while off-diagonal values indicate transfer to other months. IoU is used because it directly measures the spatial overlap between predicted and reference building masks and is less dominated by class imbalance.

The transferability matrix is characterised by a clear seasonal structure with the highest values concentrated in the central part of the matrix, specifically May, June, July, and August. This applies to both the classified images and the models. The values range from 0.39 to 0.44. In 10 of 12 acquisition-month rows, the highest median IoU occurred when the model-selection month matched the acquisition month of the classified scenes. It confirms expectations that the highest accuracy is observed when classified images were acquired in a similar time of the year as when the model was trained. For the August and September models, scenes acquired in July performed about 0.01 better than those acquired in the corresponding training month. Besides a rectangular high-accuracy region in the spring/summer months, models trained on February and March imageries, exhibit moderate yet relatively stable transferability, with median IoU values typically ranging between 0.32 and 0.41 through the period from February to October. Model trained on April imagery shows the late spring - summer (April - August) transferability. 

A distinct decline is observed for scenes acquired after September, especially when applying summer-trained models. 

The lowest performance from all classifications was observed when the winter models were used for training. The December model resulted in 0.09-0.13 accuracy for images from almost all months, except January (0.19) and December (0.17). Winter conditions introduce substantial temporal variability in land surface appearance, driven by snowfall and changes in snow cover extent and state. Under the adopted selection criteria, the best performing model for December was trained on a snow-covered scene, which suggests that snow cover enhanced the spectral contrast between buildings and their surroundings. This is also a limitation: the model performs poorly on scenes where snow conditions differ or are absent.

\subsubsection{Seasonal sensitivity across settlement types}
\label{sec:crosstemp_settlement}
The previous analysis showed that model performance depends strongly on the acquisition season. Here, we examine whether this seasonal effect is the same across all built-up environments or varies with settlement structure. This distinction is important because Sentinel-2 pixels represent dense urban blocks, industrial roofs, suburban houses, and dispersed rural buildings in very different ways. Compact built-up areas have larger, more continuous building surfaces, whereas low-density settlements often include small or sub-pixel buildings mixed with vegetation, bare soil, roads, and shadows. As a result, the same monthly model may be stable in dense urban areas but less reliable in suburban or rural settings. To evaluate this effect, the L2A Phase~2 results were grouped by the eight test regions introduced in \Cref{sec:study_area}, representing industrial, dense urban, medium-density residential, suburban, and dispersed rural built-up types.
\begin{figure}[H]
    \centering
    \includegraphics[width=1\linewidth]{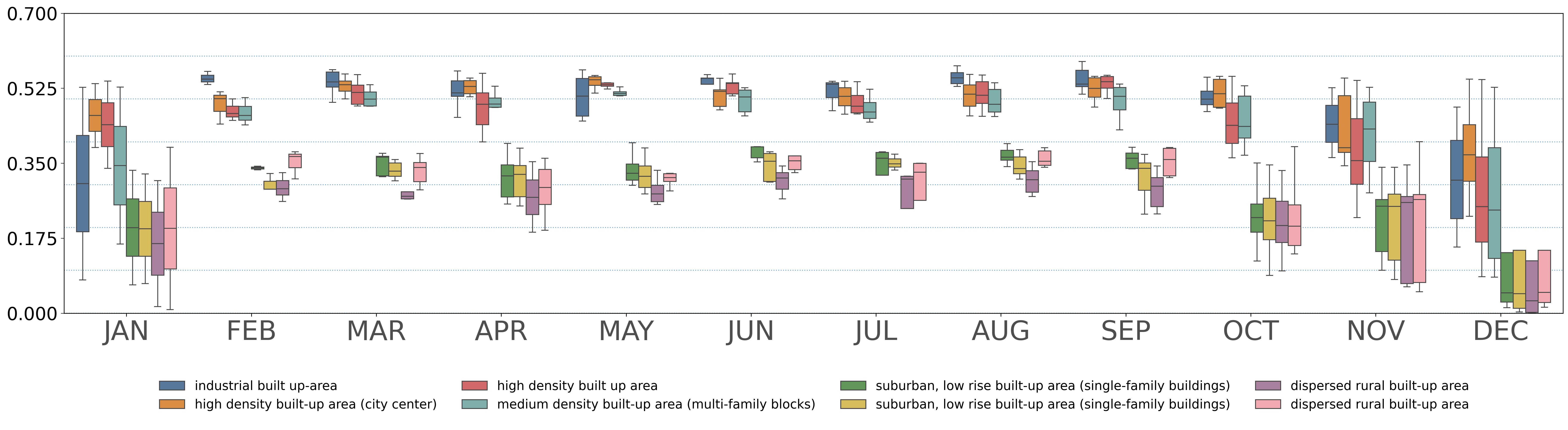}
    \caption{Monthly distribution of IoU values across the eight built-up test regions for Sentinel-2 L2A products in Phase~2. For each month, test scenes were classified using the corresponding best monthly model selected in Phase~1.}
    \label{fig:fig10_IoU_phase2}
\end{figure}

\Cref{fig:fig10_IoU_phase2} shows the accuracy levels of IoU for each month, broken down by built-up types. Detailed metrics for individual test areas (R1--R8) are provided in Appendix~\Cref{tab:metrics_regions_phase2}.
The comparison reveals a clear separation between compact and low-density built-up areas. Regions R1--R4, representing industrial zones, city centres, and dense urban development, consistently achieve higher classification accuracy (median IoU: 0.49--0.53) than regions dominated by suburban and dispersed rural structures, R5--R8 (median IoU: 0.28--0.34). The spread within both groups widens markedly in the winter months.

From February until September, the IoU values are stable and oscillate around 0.5 for R1-R4 and 0.35 for R5-R8. The highest results across regions are recorded in late summer and early autumn, especially in August and September. In contrast, late autumn and winter months (November–January) exhibit lower performance and higher dispersion for nearly all regions, confirming the strong seasonal sensitivity of cross-scene generalization. Dispersion increases earlier in dispersed built-up areas (R5–R8) than in the dense urban fabric (especially industrial zones and city centres), already rising in September and October.

In our setup, each monthly model is trained on a single scene, which can expose the model to worst-case conditions: if the training image contains ephemeral features like snow, the model may learn scene-specific features, reducing transferability to scenes with different or absent snow conditions. December represents this case. In an operational setting, where winter-season classification is required, the effect can be mitigated by training on a temporal composite (e.g. median) or by ensembling models fitted to several acquisitions from the same month, so that no individual phenomena dominates the learned representation. 

\begin{figure}[htbp]
    \centering
    \includegraphics[width=0.8\linewidth]{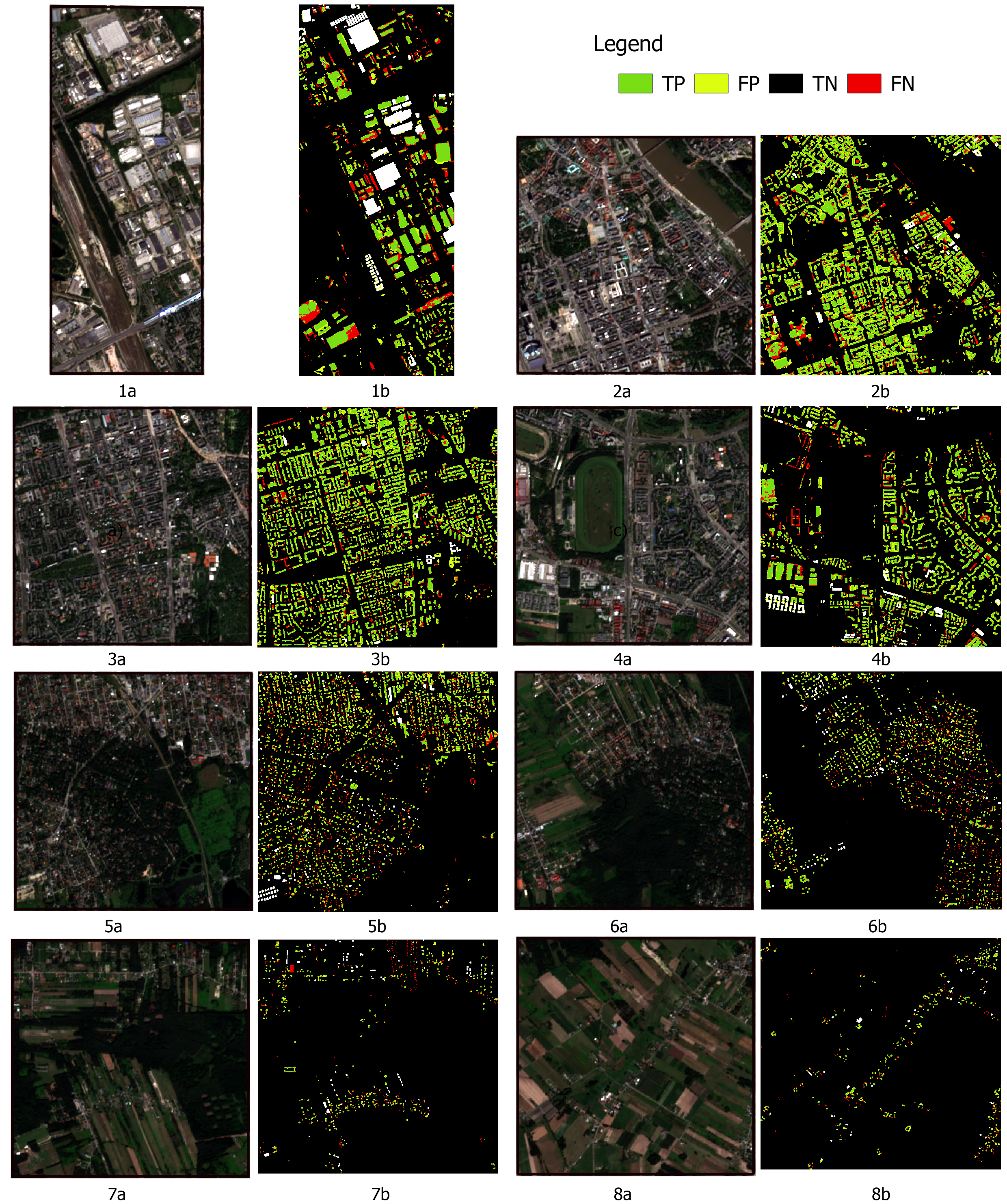}
    \caption{Results of the classification with the highest mean IoU across all areas. The regions are representative for different urban structures: R1 –industrial built-up area, R2 – high density built-up area (city centre), R3 – high density built up area, R4 – medium density built-up area (multi-family blocks), R5 and R6 – suburban, low rise built-up area (single-family buildings), R7 and R8 – dispersed rural built-up area. Sentinel-2 imagery - a,  classification results - b .}
    
    \label{fig:8x3_zblizenia}
\end{figure}

The \Cref{fig:8x3_zblizenia} shows the example of classification map for analysed areas with the highest mean IoU value across all areas (0.47) - based in imagery acquired on 12 August 2023. 
Classification quality varies systematically with built-up typology. In compact industrial and high-density urban areas (R1–R4), true positives (green) dominate building footprints, with false positives (yellow) clustered mostly along edges. False negatives appear both for large industrial halls with uniform roof surfaces and for small buildings that are not distinguished from the background. The dense urban fabric is reproduced almost completely, with errors limited to courtyards and narrow inner-block spaces.

In low-density suburban and rural scenes (R5–R8), the proportion of correctly detected buildings drops sharply. Scattered single-family houses produce only sparse green pixels against extensive correctly classified background, and a substantial share of individual structures is missed entirely. 

Additional examples illustrating typical successes and failure modes, including large warehouse halls, dense historic urban fabric, high-rise buildings affected by seasonal shadows, glass canopies, and greenhouse-like structures, are provided in~\ref{app:classification-examples}.

\subsubsection{Quantitative area estimation}
\label{sec:crosstemp_quantitative}
\begin{figure}[H]
    \centering
    \includegraphics[width=1\linewidth]{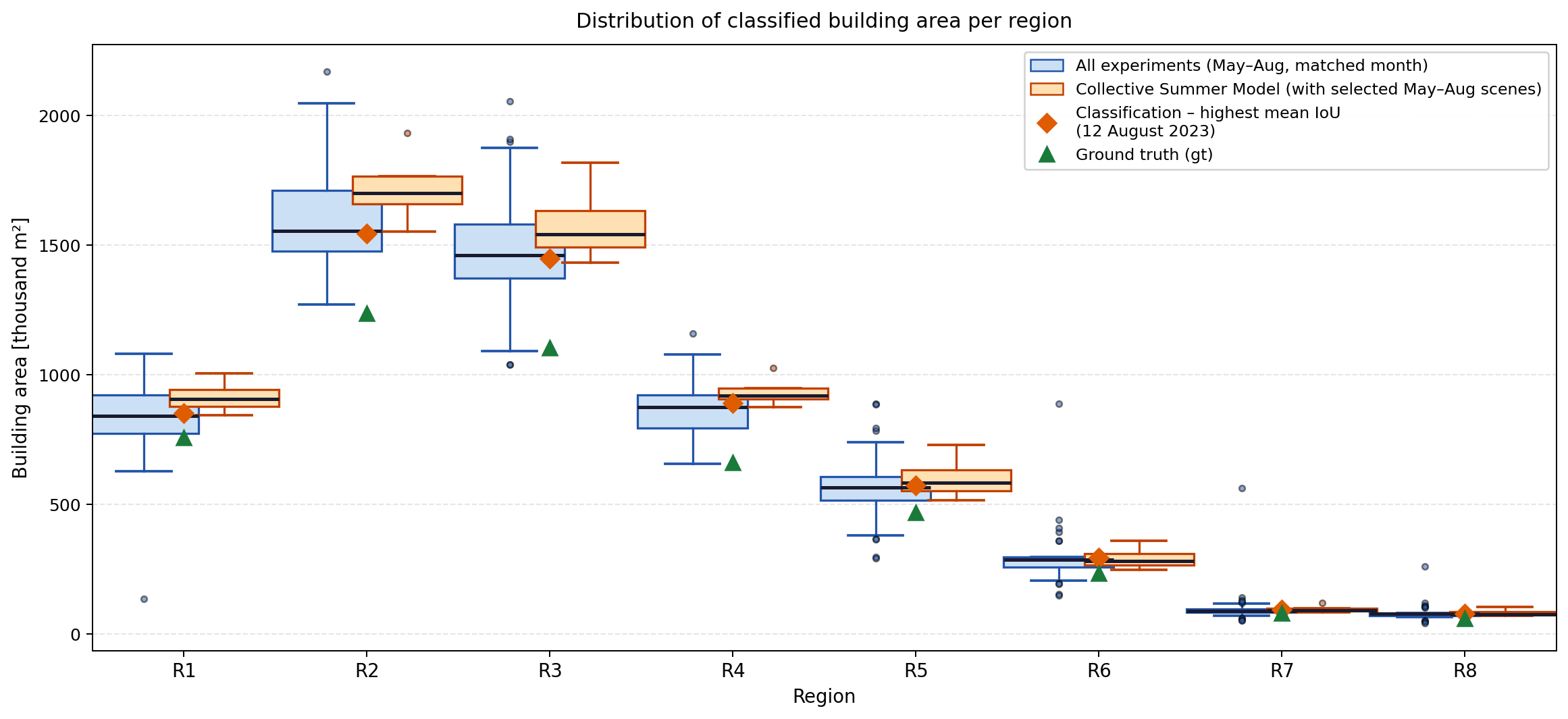}
    \caption{Distribution of classified building area per region, based on the matched-month checkpoints from best period of year (May–September), one best classification and the Collective Summer Model (see \Cref{sec:seasonal_transferability}). R1 –industrial built-up area, R2 – high density built-up area (city centre), R3 – high density built up area, R4 – medium density built-up area (multi-family blocks), R5 and R6 – suburban, low rise built-up area (single-family buildings), R7 and R8 – dispersed rural built-up area.}
    \label{fig:phase3_unet_area_boxplot_plus4month}
\end{figure}

For many practical applications, the total estimated built-up area is as important as the spatial accuracy of individual building pixels. Urban planning, population estimation, land-use monitoring, and historical change analysis often rely on aggregate built-up surface estimates rather than on pixel-wise classification maps alone. We therefore assessed whether the models systematically over- or underestimate total building area. Predicted building area was computed as the total area of pixels classified as buildings and compared with the reference area derived from BDOT10k. We used classifications from the best-performing period (May--August), the August model trained on imagery acquired on 12 August 2023, and the Collective Summer Model -- a model fine-tuned jointly on four selected L2A scenes, one from each month of the stable May--August period, to reduce dependence on a single acquisition date (see \Cref{sec:seasonal_transferability}). Results are shown in \Cref{fig:phase3_unet_area_boxplot_plus4month}.

Across all regions, the predicted building area exceeds the BDOT10k reference area, indicating a systematic positive bias. In most regions, the reference value lies below the lower quartile of the predicted-area distribution, showing that the overestimation is not limited to individual scenes. This pattern is consistent with false-positive detections near building boundaries at Sentinel-2 resolution, where mixed pixels along roof edges may be assigned to the building class (see \Cref{sec:loc_error}). The bias occurs across all settlement types, but its character differs by region: the largest absolute overestimation is observed in the city-centre and high-density built-up areas (R2--R3), whereas the largest relative bias occurs in dispersed rural areas (R7--R8), where the true built-up area is small. The closest agreement with the reference data is observed in the suburban low-rise areas (R5--R6).

\subsection{Separating detection from localisation error}
\label{sec:loc_error}
We examined how much of the observed pixel-wise error is due to missed or incorrect building detection, and how much is concentrated near building boundaries. In the latter case, disagreement is expected because of the mismatch between \(10\,\mathrm{m}\) Sentinel-2 imagery, much finer BDOT reference data, and possible small shifts introduced by resampling. Here, boundary error is interpreted primarily as a localisation problem: the model may correctly identify the presence of a building but place its footprint slightly too far inward or outward, or shift it relative to the reference. To address this, we complement the global evaluation with a simple morphology-based analysis that separates stable building interiors and stable background areas from the immediate boundary neighbourhood. This allows us to distinguish detection quality from boundary-localisation error and assess whether the residual disagreement reflects missed buildings or primarily reflects uncertainty in footprint placement along building edges.

This analysis was carried out on a selected random image chosen for detailed inspection of the boundary effect. Under the standard full-image evaluation, the model achieves \(BA=0.7998\) and \(F1=0.6770\). When the evaluation is restricted to the stable building core and far background, the score increases to \(BA_{\mathrm{guard}}=0.8318\) and \(F1=0.7190\). This indicates that significant part of the disagreement observed under the standard protocol is concentrated in the immediate vicinity of building boundaries, rather than in the semantically stable regions of the scene (i.e. building interior and background). The boundary-specific metrics confirm this interpretation. Boundary recall in the inner one-pixel band is only \(R_{\mathrm{bnd}}=0.4381\), substantially lower than the corresponding core recall of \(0.6978\). This points to a marked drop in performance from the interior of buildings to their edges. On the negative side, boundary specificity reaches \(S_{\mathrm{bnd}}=0.7126\), whereas specificity in the far background is much higher at \(0.9659\), showing that false positives occur much more frequently close to building outlines than in further background. Consistently, the boundary error share equals \(0.2974\), meaning that estimated \(30\%\) of all raw errors are concentrated within the narrow \(\pm 1\) px boundary zone. Taken together, these results indicate that the significant amount of classification error is linked to boundary localisation rather than failure to recognize building presence in general.

We find the observed effect consistent with the expected behavior of a medium-resolution model evaluated against a much finer reference source. Building interiors are detected substantially more reliably than building edges, while false positives remain limited in clear background but increase in the immediate surroundings of the reference footprint. This indicates that a considerable part of the remaining disagreement is not due to complete failure to recognize buildings, but rather to imperfect localisation of their boundaries. In the context of the paper, this finding refines the interpretation of the main quantitative results: the standard metrics remain valid as measures of strict pixel-wise agreement, but they are somewhat conservative with respect to semantic building detection. The morphology-based analysis therefore suggests that the model captures building presence more robustly than would be inferred from the raw confusion matrix alone, and that a substantial share of the residual error is concentrated in a narrow boundary zone where uncertainty in footprint placement is most pronounced.

\subsection{Spatial transferability}
\label{spatial_transferability}
\begin{figure}[H]
    \centering
    \includegraphics[width=1\linewidth]{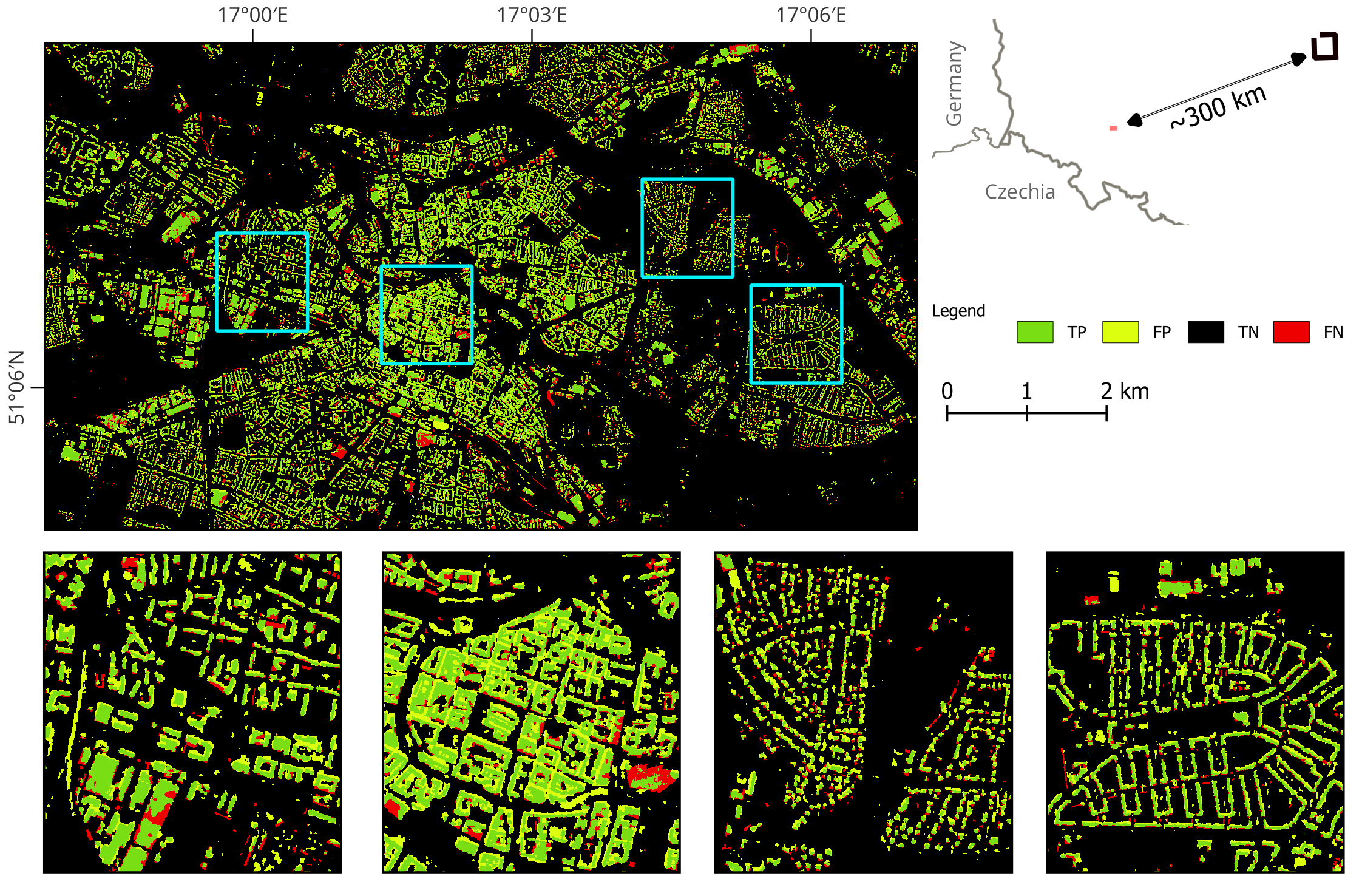}
    \caption{Classification result for the spatially distinct Wrocław test area, located outside the Warsaw training and evaluation region. The panels show selected examples of built-up structures: a medium-density residential area with multi-family blocks, high-density city-centre development, suburban low-rise single-family housing, and another medium-density residential area.}
    \label{fig:wroclaw}
\end{figure}

To assess model behaviour beyond the analysed region, we additionally classified the city of Wrocław and its surroundings (Sentinel-2 tile 33UXS). This area differs from the training region because of its distinct historical development pattern, different distribution of building types and layouts, and the milder climate of south-western Poland, which may produce different seasonal vegetation and surface conditions in the imagery. The model generalised well to this spatially distinct area, achieving IoU=0.468, F1=0.638, and BA=0.833, which is comparable to the values obtained in the original test region. The results are presented in \Cref{fig:wroclaw}.

\section{Discussion -- operational guidelines for Sentinel-2 building mapping}

The study presented a comprehensive analysis of how built‑up areas can be classified using Sentinel‑2 images and deep learning methods. A classification model architecture was examined, along with the recommended level of satellite data processing, temporal transferability and building typology as factors influencing the detection performance. 

\subsection{Model comparison}
The first part of the study analysed the built-up maps produced by the U‑Net and DeepLabV3+ models. Both models were trained using Sentinel‑2 data at processing levels L1C and L2A, separately. The BDOT10k, a widely available spatial vector reference dataset containing among other things outlines of buildings in Poland, served as the ground‑truth data source for buildings locations.

The comparison of classification results showed that the U‑Net performed better than DeepLabV3+. In both cases, classification errors were mainly caused by the spatial resolution of Sentinel‑2 data. U‑Net identified building boundaries more precisely, including narrow buildings and inner courtyards in the city centre. The U‑Net results also had a smaller IQR range, indicating greater stability. This finding is consistent with results reported across multiple independent studies on Sentinel-2 building extraction. Dixit et al. \cite{dixit2021dilated} compared seven architectures on Sentinel-2 data from three Indian cities and found that a U-Net variant achieved the highest F1-score (0.47) and mean IoU (0.58), with standard U-Net ranking second (0.42 and 0.57, respectively). Feng et al. \cite{fengNationalscaleMappingBuilding2023} tested six super-resolution semantic segmentation network variants on real Sentinel-2 imagery across Chinese cities and found that NASUnet-based architectures (U-Net variants with a NASNet-Mobile backbone) consistently outperformed both HRNet-based and DeepLab-based alternatives within the same framework.

\subsection{Output resolution}
Both models produce output at approximately 2.8 m resolution, almost four times finer than the native 10 m Sentinel-2 data. This allows individual buildings and courtyards to be distinguishable in dense urban areas, and improves the detection of smaller buildings. The resolution increase comes from our patch resampling strategy: 64-pixel patches covering 640 × 640 m are resampled to 224 pixels to match the pretrained backbone input size. This is not a dedicated super-resolution (SR) module. This distinguishes our approach from recent end-to-end super-resolution semantic segmentation (SRSS) methods. Xu et al. \cite{xuESPC_NASUnetEndtoEndSuperResolution2021} used an efficient subpixel convolution module before the segmentation encoder. Feng et al. \cite{fengNationalscaleMappingBuilding2023} extended this with a deeper Enhanced Deep Super-Resolution (EDSR) module. Both approaches treat SR as a distinct learned operation within the network. Our implicit approach, relying on the U-Net decoder's inherent spatial refinement through skip connections and successive upsampling, produces results comparable to those dedicated SRSS methods. Feng et al. \cite{fengNationalscaleMappingBuilding2023} reported IoU values of 0.311–0.499 across four Chinese test cities using their EDSR\_NASUnet on real Sentinel-2 imagery, with a mean around 0.375. Our peak IoU of approximately 0.43 falls within this range, achieved with a substantially simpler architecture. This is consistent with Xu et al.'s \cite{xuESPC_NASUnetEndtoEndSuperResolution2021} finding that more complex SR modules (Residual Dense Super-Resolution - RDSR, based on Residual Dense Networks) actually performed worse than the simpler Efficient Sub-Pixel Convolution (ESPC) module when integrated into SRSS networks, and that excessive depth in the feature extraction component can destabilise gradient propagation. Our results extend this observation: the encoder-decoder's inherent upsampling capacity with data deliberately selected in terms of image acquisition dates without any dedicated SR module may be sufficient for Sentinel-2 building extraction. 

Increasing the resolution is also possible by applying an SR model before classification, as a stagewise approach. This requires an appropriate SR model and higher‑resolution images in case additional training is needed. However, high‑resolution data usually contain fewer spectral bands than Sentinel‑2, meaning that only selected bands can be enhanced, which may reduce classification performance. Depending on the SR model, distortions of the original spectral data may also occur. Xu et al. [14] compared stagewise and end-to-end SRSS approaches and found that end-to-end methods were less affected by the quality degradation of low-resolution input images, as the two tasks can jointly optimise for building-relevant features rather than general image reconstruction.

\subsection{Processing level of Sentinel-2}
Sentinel-2 data are distributed as L1C (top-of-atmosphere reflectance) and L2A (atmospherically corrected surface reflectance). L2A is commonly used in environmental studies to reduce atmospheric variability across acquisition dates.

The U-Net model was pretrained jointly on L1C and L2A data ~\cite{wangDecouplingCommonUnique2024a}, then fine-tuned separately at each processing level using the same scenes acquired at both processing levels. L2A produced slightly higher accuracy overall: median IoU of 0.369 versus 0.359 and median F1-score of 0.535 versus 0.525, an improvement of roughly 2–3\%. However, this advantage was not consistent across months. L2A outperformed L1C in six months, including January, February, April, and the summer period. In the remaining six months — including May, June and the late-autumn/winter months of November–December — L1C performed comparably or slightly better.

Most comparable studies used only one processing level without comparison: Dixit et al. \cite{dixit2021dilated} used L2A exclusively,  while Sirko et al. \cite{sirkoContinentalScaleBuildingDetection2021} and Feng et.al \cite{fengNationalscaleMappingBuilding2023} used L1C exclusively. The absence of comparative L1C/L2A analyses in the building extraction literature makes our finding, notably that the difference is marginal, a useful practical contribution. 

\subsection{Seasonal Transferability and Temporal Patterns}
\label{sec:seasonal_transferability}
\begin{figure}[h!]
    \centering
    \includegraphics[width=1\linewidth]{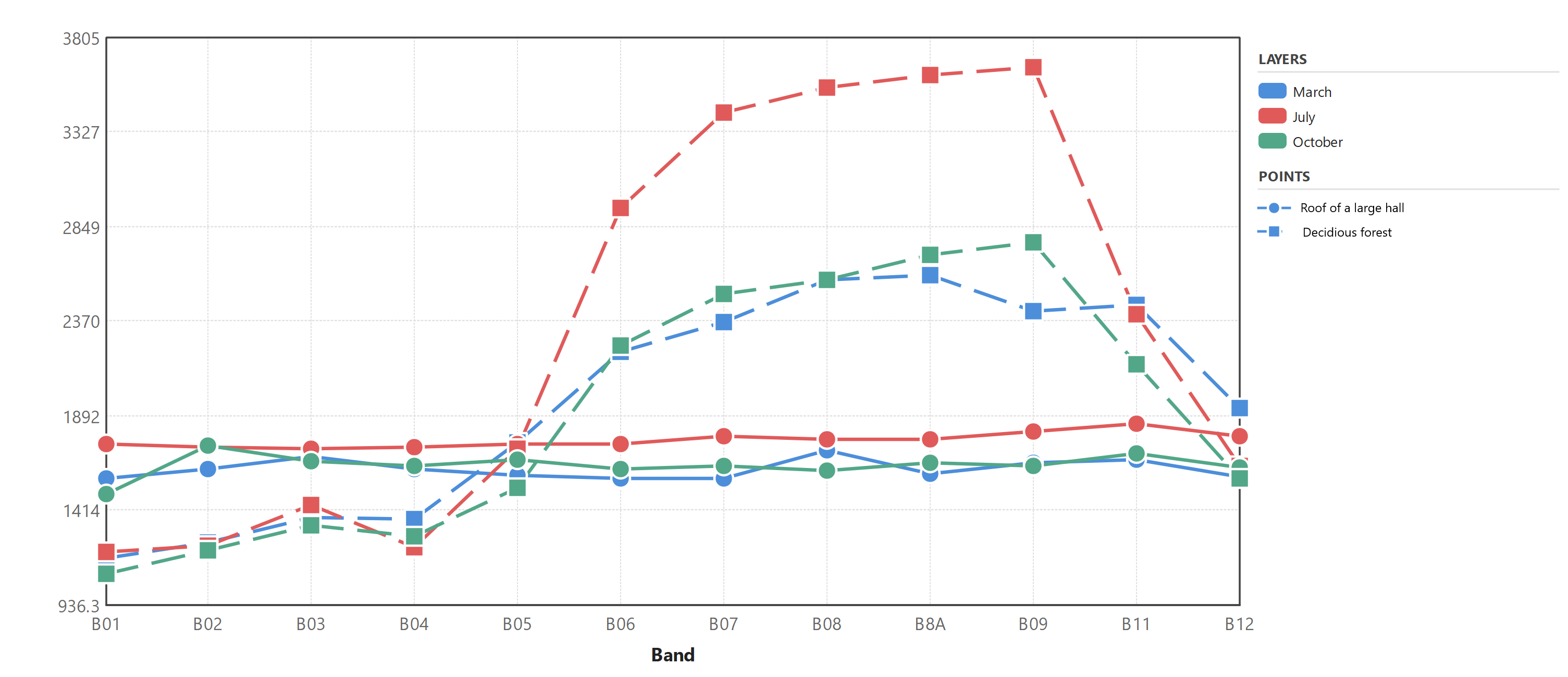}
    \caption{Comparison of spectral curves of a roof and deciduous forest recorded in the Sentinel-2 images in the months of March, July and October.} 
    \label{fig:spectral}
\end{figure}

\begin{figure}[h!]
    \centering
    \includegraphics[width=1\linewidth]{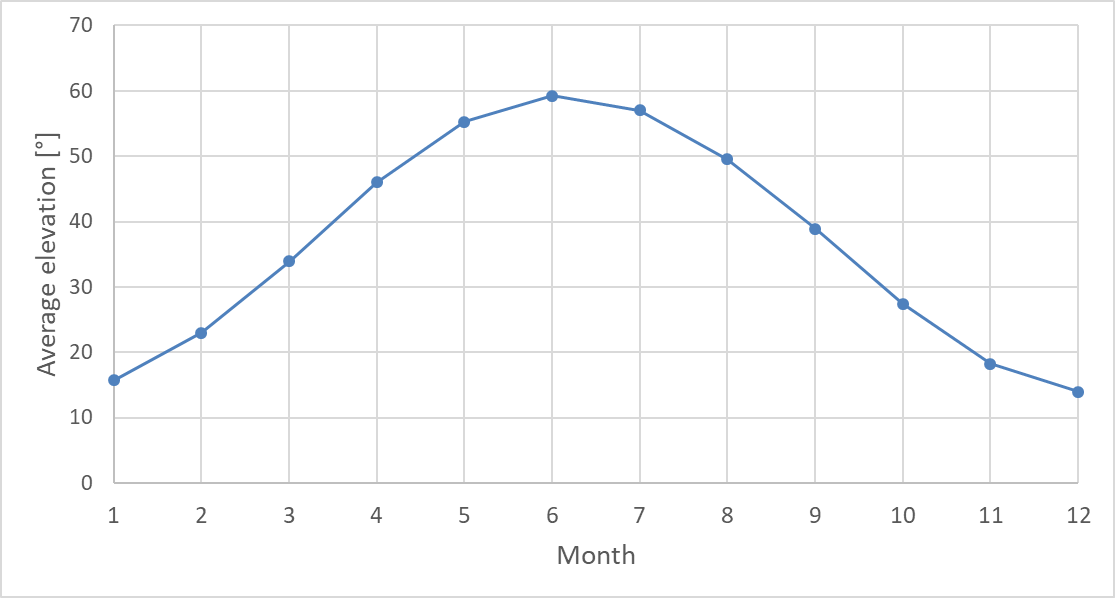}
    \caption{The sun elevation for Warsaw at 9:45 UTC (during Sentinel-2 acquisitions).} 
    \label{fig:sun_angle_monthly}
\end{figure}

\begin{figure}[h!]
    \centering
    \includegraphics[width=0.48\linewidth]{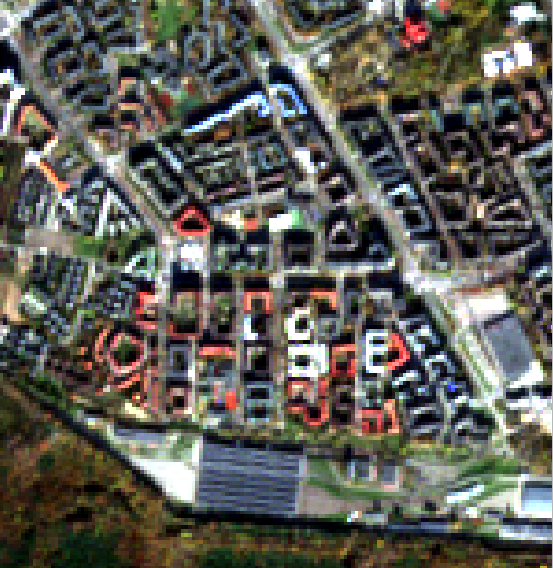}\hfill
    \includegraphics[width=0.48\linewidth]{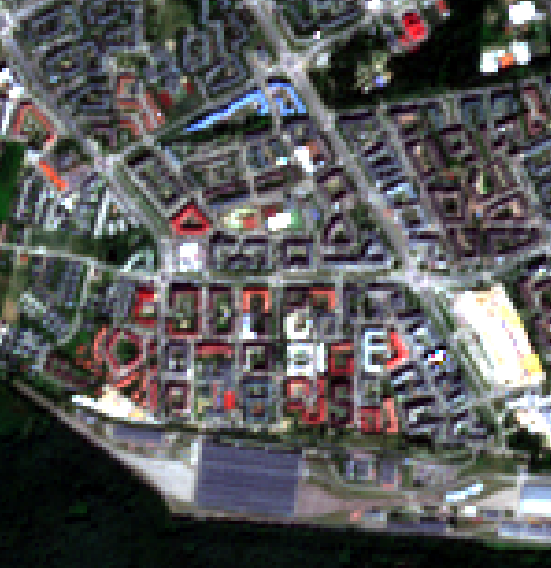}
    \caption{Comparison of building shadows at Sentinel-2 images recorded at the end of October and July}
    \label{fig:shadows}
\end{figure}

Within \Cref{fig:fig_IoU_transferability_heatmap_phase2}, a distinct block of high IoU values is visible for models trained and applied on data from May, June, July, and August. Any combination of model and image from these months resulted in high performance. Outside this period, particularly in late autumn and winter, accuracy decreased,  even when the training and inference months were the same (the main diagonal of \Cref{fig:fig_IoU_transferability_heatmap_phase2}). This confirms that seasonal conditions affect classification performance regardless of whether the model was specifically trained for that period.
The models confirm the well-known principles used in the traditional visual interpretation of satellite data. Hence, the interpretation of land cover classes is based on images taken during intensive vegetation periods, which ensures the greatest distinguishability of classes. In the case of buildings, vegetation serves as a high-contrast background, helpful for building recognition. As an example, \Cref{fig:spectral} shows the spectral curves of the roof and deciduous forest recorded in the Sentinel-2 images in March, July and October. The differences in the roof's spectral reflections across individual months are not large. Deciduous forest, on the other hand, is characterised by a large increase in values in channels 6, 7, 8, 8A and 9. This significantly increases the ability to distinguish between buildings and forests. The same is true for other vegetation classes. An additional factor affecting the recognition of buildings is the sun's elevation. \Cref{fig:sun_angle_monthly} shows a chart of the sun elevation for Warsaw over a year~\footnote{Sun elevation angle was estimated with the NOAA calculator \url{https://gml.noaa.gov/grad/solcalc}}. In May, June, July and August, the Sun is high, so the shadows are relatively small and do not impede building recognition. An example illustrating the shadows of buildings in October and July is shown in \Cref{fig:shadows}.

Vegetation conditions should also be considered when interpreting seasonal transferability, because vegetation is often the dominant background around buildings. Its spectral response varies with phenology and weather conditions, and the timing of these changes may vary from year to year. As a result, a model selected for a given month may perform best when applied to imagery acquired under similar seasonal conditions, and not necessarily to the same calendar month in every year. To reduce dependence on a single acquisition date, we additionally evaluate a Collective Summer Model (CSM), fine-tuned jointly on four L2A scenes from the stable May--August period, one scene per month.
\begin{figure}[h!]
    \centering
    \includegraphics[width=1\linewidth]{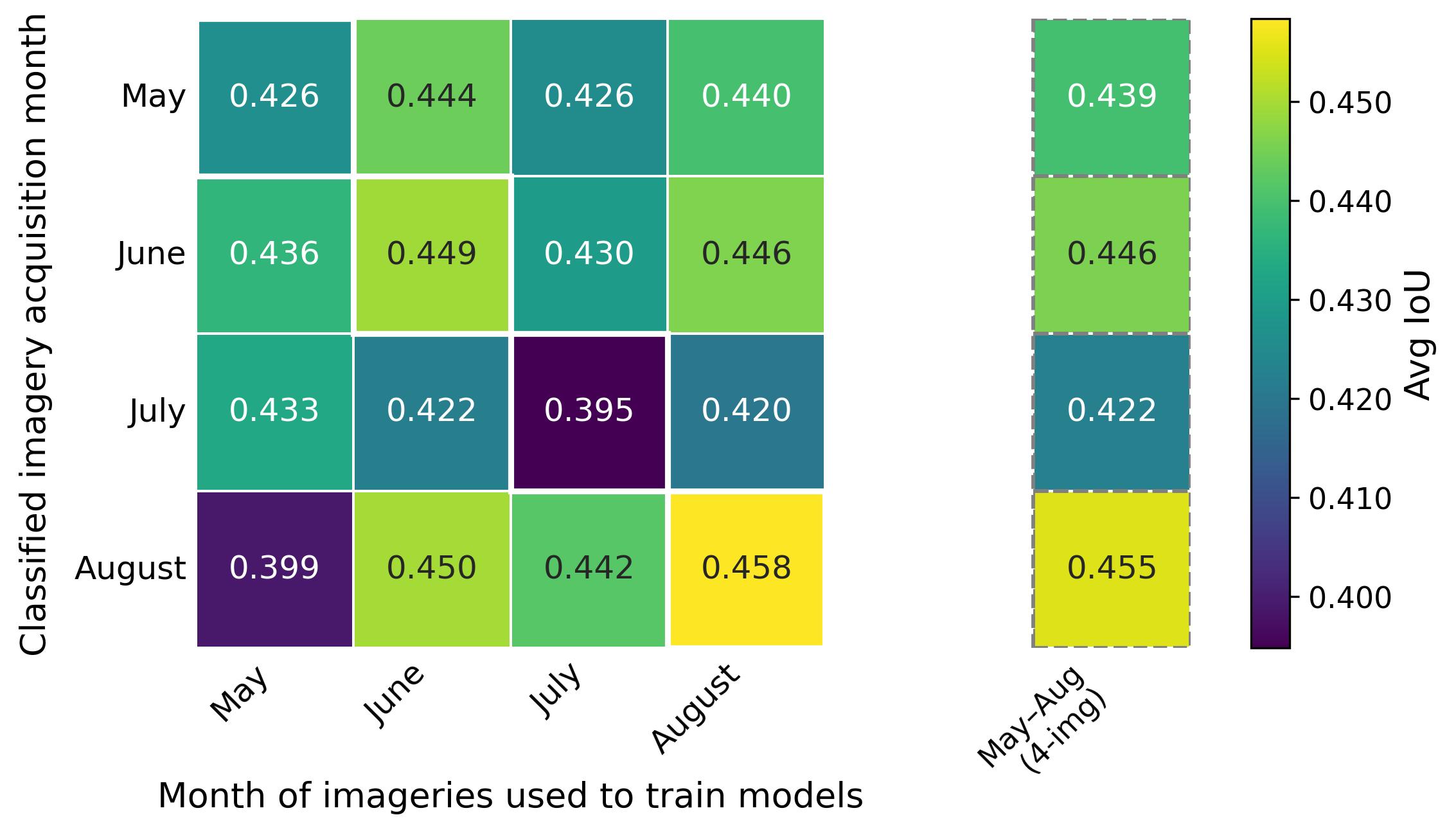}
    \caption{Comparison of single monthly models and the Collective Summer Model (CSM) for the stable May--August period. Four independent test scenes, one from each month, were classified using the corresponding single-month models (first four columns) and the CSM fine-tuned jointly on May--August scenes (last column). Values show median IoUs across the eight test regions.}
    \label{fig:iou_heatmap}
\end{figure}
\Cref{fig:iou_heatmap} compares the single monthly models with the CSM on independent May--August scenes. The CSM achieves performance similar to, and in some cases slightly better than, the month-specific models. This indicates that combining representative scenes from the stable summer period can reduce dependence on a single acquisition date while preserving the accuracy obtained with specialised monthly models. It therefore provides a practical option for limiting local and inter-annual variability in operational classifications.

\subsection{Building characteristics}
\Cref{fig:fig10_IoU_phase2} shows IoU distributions across building typology categories for each month. Industrial and high-density built-up areas, including city centre blocks and high-density residential, consistently achieved the highest median IoU values, remaining relatively stable from February to September (typically achieved IoU>0.525 in favorable months). These typologies are characterised by large, contiguous rooftop surfaces and high building coverage ratios, which produce a stronger and more consistent spectral signal at 10 m resolution.

The largest and most consistent accuracy drop was observed in suburban and dispersed rural typologies -- areas dominated by single-family houses and scattered buildings. These categories showed the lowest median IoU values across all months, and the widest IQR, particularly in Autumn and Winter (typically achieving IoU values of ~0.350 in favorable months). This pattern reflects the fundamental challenge of detecting small, isolated buildings at 10 m resolution: individual buildings may occupy less than one pixel, making them highly sensitive to shadow, surrounding vegetation, and positional misalignment with the reference data.

The seasonal decline in accuracy discussed in \Cref{sec:seasonal_transferability} is therefore not uniform across the scenes -- it is concentrated in low-density typologies. High-density areas remain relatively well detected even in autumn, while in sparse settlement areas seasonal degradation is most visible.

\subsection{Limitations and future work}
While the presented recommendations are intended to be practical for routine Sentinel-2 building mapping, they should be interpreted as operational guidance within the scope of this study rather than universal rules. First, our evaluation is geographically constrained to the Warsaw region and its surroundings in central Poland, covering a specific mix of settlement morphologies, land cover, and climate. Consequently, the observed optimal months, stability windows, and cross-season transfer patterns are expected to be most reliable for comparable conditions, and their direct transfer to other regions should be treated with caution. Extending the analysis to different climatic zones, vegetation regimes, and urban fabrics (e.g., Mediterranean, boreal, arid, mountainous, and coastal settings) is therefore a key direction for future work. Second, several of the seasonality-related explanations -- such as the potential role of low solar elevation, shadowing, and intermittent snow cover in winter months -- are plausible but were not explicitly isolated and validated in controlled experiments. Future studies should quantify these factors more directly, for example by stratifying scenes by solar geometry, acquisition time, and snow/ice indicators, and by testing whether similar effects appear under comparable illumination conditions outside the nominal winter period. Third, from a modeling perspective, we benchmark only two convolutional segmentation architectures (U-Net and DeepLabV3+). Although this choice provides a stable and reproducible basis for analyzing spatio-temporal effects, a broader benchmark including more recent model families (e.g., transformer-based or hybrid architectures) would help determine how architecture choice interacts with seasonality, settlement type, and the detection of small or sub-pixel buildings. Finally, future work should examine whether the derived month-wise model-selection strategy remains robust under different training set sizes, labeling sources, and operational constraints, and whether similar guidelines can be learned or transferred across regions using domain adaptation or multi-region training.

\section{Conclusions}
Monitoring built‑up areas is an important part of environmental and economic activities. The construction time of a new building depends on many factors, but it can be assumed that information about built-up areas should be updated once a year. With this in mind, a good classification strategy is to create a model that can be used every year.

This research demonstrates that seasonal variations significantly influence the performance of deep-learning models for building detection from Sentinel-2 imagery. This confirms the principles of image selection used in visual interpretation and traditional classification methods. Moreover, the study reveals insights for practical application of robust DL models across different temporal and spatial contexts related to building detection.

First, L1C (top-of-atmosphere) products and L2A (atmospherically corrected) products provide very similar results across all evaluation metrics and months, with a slight advantage of L2A. The May to August evaluations showed particularly strong performance, suggesting that these months may represent optimal imaging conditions in the study region.

Second, pronounced seasonal degradation occurs during transition seasons and especially in winter months. This loss correlates with winter conditions when low sun angles introducing bigger shadows and spectral responses related to vegetation are significantly slowed down. 

Third, the cross-scene classification results (Phase 2) reveal that individual monthly models trained on imagery from a specific month generalize better to scenes acquired in that same month than to scenes from other months. This suggests that the image acquisition conditions related to seasonal variations have an impact on the training data, and can limit model transferability across seasons.

Fourth, performance variability across different built-up area types (\Cref{fig:fig10_IoU_phase2}) indicates that built-up classification is not uniform. Industrial areas and high-density built-up zones typically achieved better results compared to dispersed built-up areas and single-family residential zones. The classification results are influenced by the size of objects.

Fifth, the analyses showed that the recommended approach for classifying built‑up areas on Sentinel‑2 images is to use models created for four months: May, June, July, and August. Creation of a Collective Summer Model, reduces differences in vegetation phenology. This model can be used in the following years without the need for retraining. It also allows the collection of cloud‑free images, which may be difficult to obtain within a single month.

Sixth, analysis of models performance clearly indicates that general models (one model for all types of buildings) perform better in areas with higher density of buildings. Models dedicated to certain types of built-up density could reach better performance. However, such a procedure requires knowledge of the boundaries of zones of different types of development, which is usually a difficult condition to meet.

Recommendations of operational application of built-up models can be summarized:

\begin{enumerate}
    \item In the building classification task from Sentinel-2 imagery, U-Net produces better results than DeepLabV3+ (see \Cref{sec:results_phase_1_model}).
    
    \item Classify Sentinel-2 data at the processing level L2A to ensure the classification stability when using different acquisition dates (see \Cref{sec:results_phase_1_accuracy}). 
    
    \item Implement the Collective Summer Model to infer the May - August imagery to minimise the spatio-temporal differences in phenology, ensure long fine-tuned model usability and increase the chance to find high quality, cloudless imagery every year (see \Cref{sec:seasonal_transferability}). 
    
    \item If the Collective Summer Model is not available, month-specific models can also yield high-quality results. In this case, the optimal scenario assumes a library of 12 models covering every month of the year. Where only a limited model library is available, the August model may be used as a substitute for all summer months (see \Cref{sec:crosstemp_seasonal}). 

    \item Avoid training on a single acquisition in months with high temporal variance or potential domain shift (e.g. winter months with variable snow cover), as the model may overfit to transient surface conditions and fail to generalise (\Cref{sec:crosstemp_seasonal}). In such cases, training on a temporal composite (e.g. median) or by ensembling models fitted to several acquisitions from the same month is recommended (see \Cref{sec:crosstemp_settlement}).
\end{enumerate}

\section*{Acknowledgments}
The authors acknowledge the use of AI-based tools, such as ChatGPT and Grammarly, for assistance in editing, grammar enhancement, and spelling checks during the preparation of this manuscript.

\bibliographystyle{elsarticle-num}
\bibliography{CBK_IITiS}

\appendix
\section{Examples of classification results}
\label{app:classification-examples}

This appendix presents selected examples of classification outputs used for visual inspection of model behaviour.

\newcommand{\classificationexample}[3]{
\begin{figure}[htbp]
    \centering
    \includegraphics[width=0.8\textwidth]{#1}
    \caption*{\textbf{#2.} #3}
\end{figure}
}

\classificationexample
{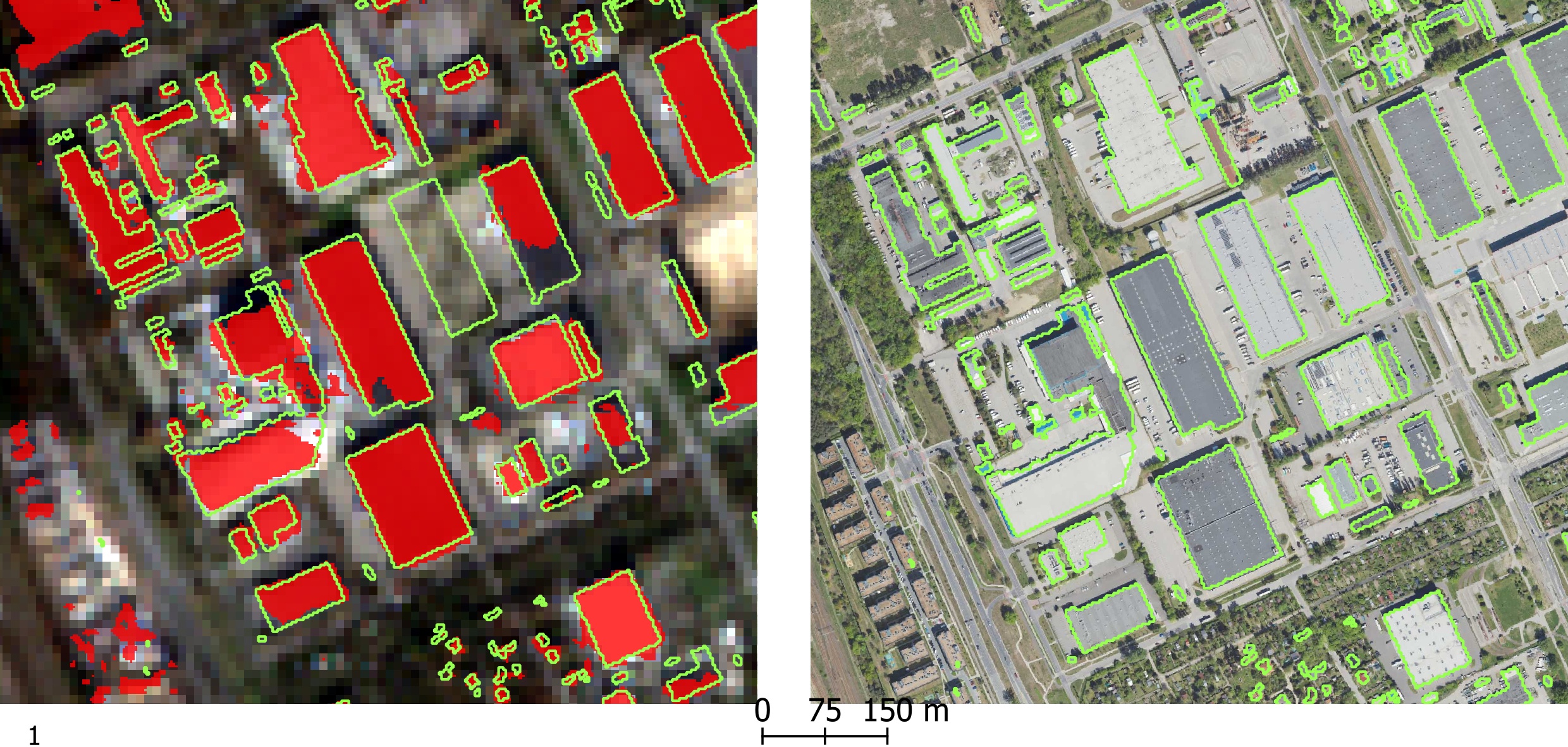}
{Warehouse halls -- high-confidence classification example}
{This example shows one of the best classifications of large warehouse halls, based on the image from 12 August 2023. Such objects are relatively easy to detect in 10 m Sentinel-2 imagery because of their large footprint, and most detections are visually consistent with the reference. However, confusion with surrounding paved surfaces remains possible, especially for simpler methods such as Random Forest, which may not clearly separate warehouse roofs from adjacent impervious areas. Reliable use of classification approach should follow the paper’s recommendations on image timing and seasonality to reduce detection uncertainty.  The high detectability of warehouse halls, combined with their frequent construction, modification, and demolition, supports the practical use of the proposed approach for e.g. monitoring warehouse development, verifying construction permits, and supporting environmental impact assessment.}

\classificationexample
{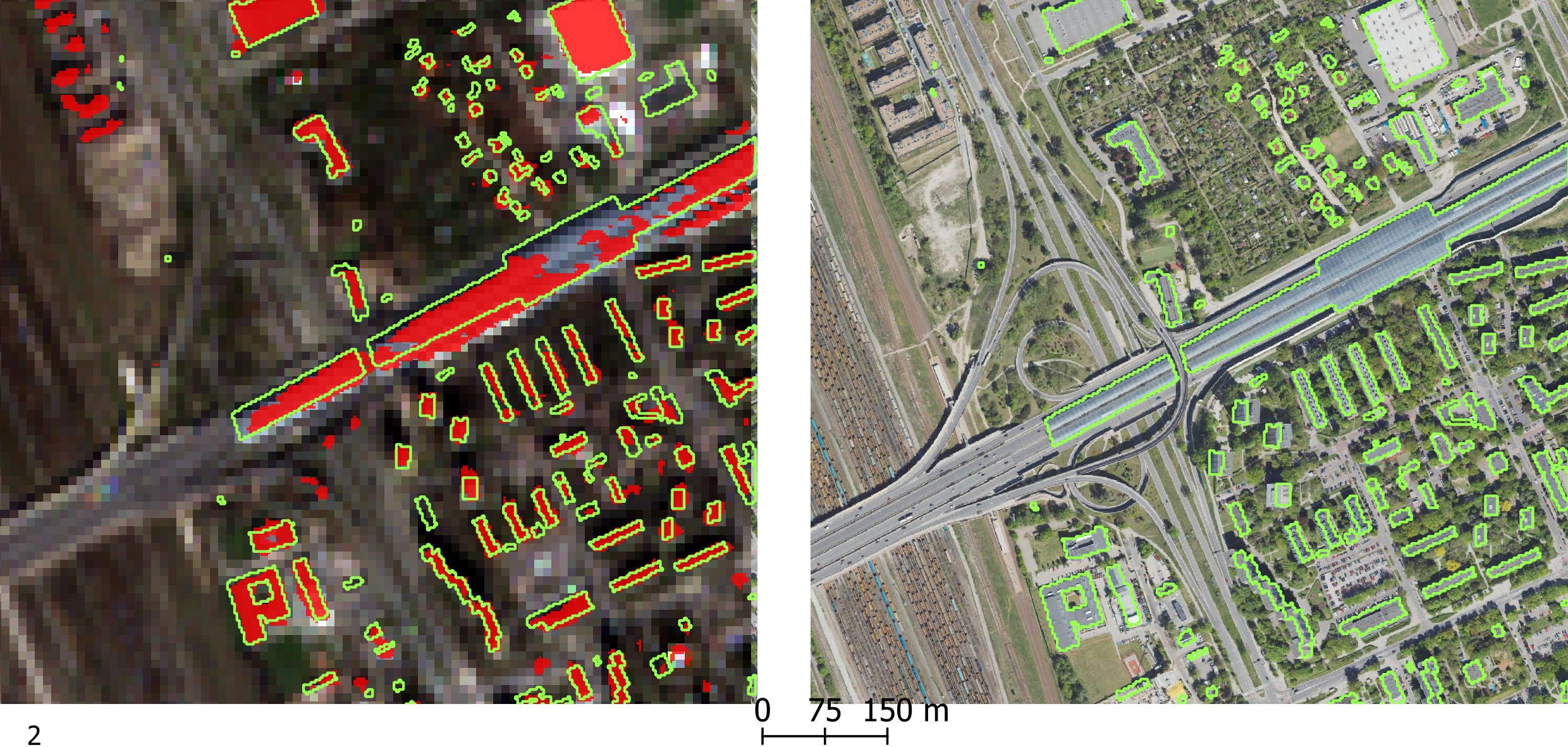}
{Glass road canopy -- selective detection of built structure}
{This example shows the detection of a glass canopy above a road, based on the image from 9 March 2024. The classification varies spatially, with frequent gaps in the detected object, likely due to changing illumination conditions and the reflective/transmissive character of the roof material. At the same time, the road surface and elements of the road junction are only minimally classified as buildings. This is a positive result: despite local false negatives within the canopy, the algorithm largely distinguishes this structure from surrounding transport infrastructure.}

\classificationexample
{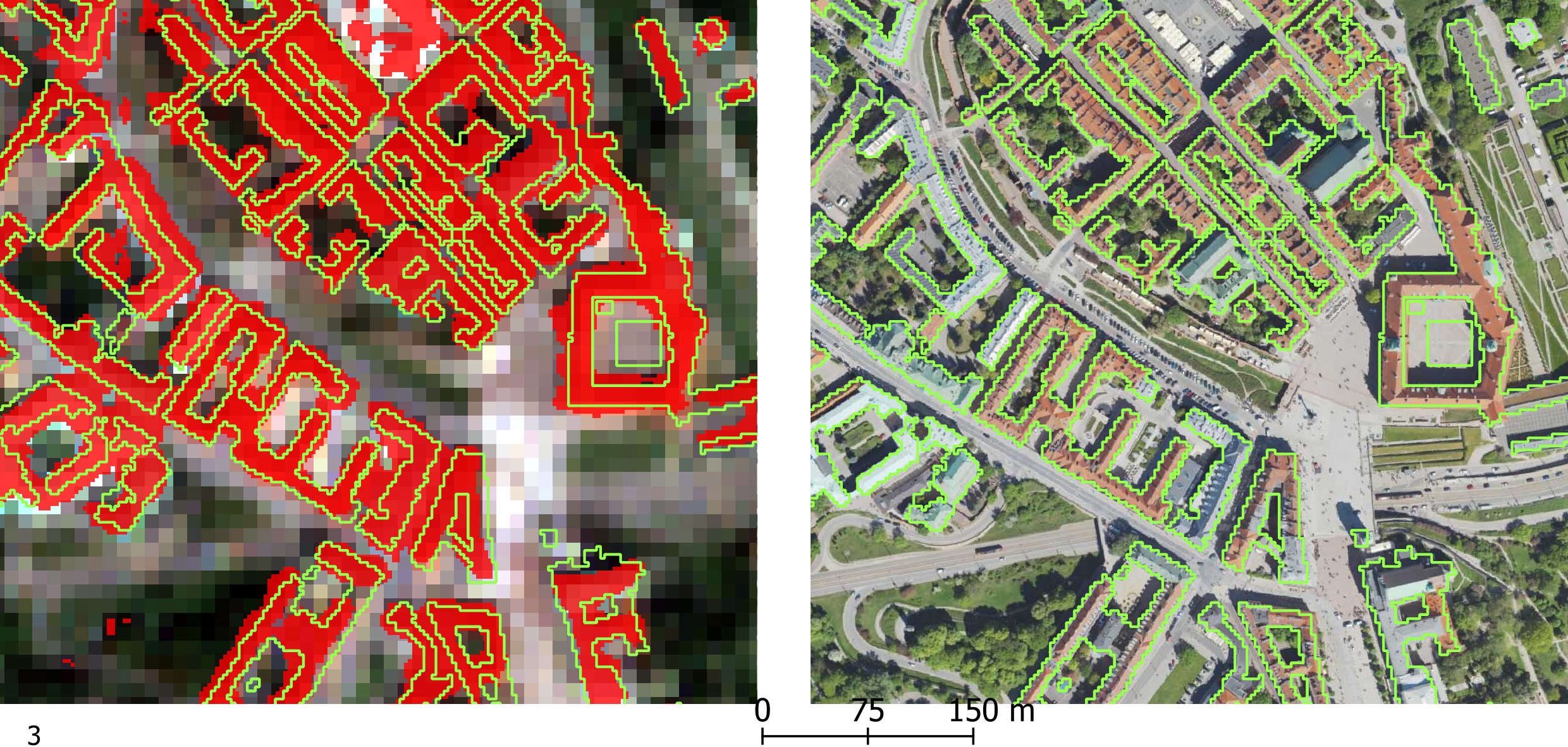}
{Dense historic city centre -- detailed classification of compact urban area}
{This example shows a good classification of dense urban development in the city centre, near the Royal Castle, based on the image from 12 August 2023. The model broadly reconstructs the shape of buildings and even preserves internal courtyards and gaps between structures. This level of spatial detail would be difficult to obtain with traditional methods such as Random Forest, especially given the 10 m Sentinel-2 input resolution and ordinary resampling without super-resolution. The result suggests that the model captures not only the presence of compact buildings, but also their approximate spatial arrangement; restaurant gardens in the market square are also detected, illustrating a remaining source of possible confusion with non-building urban objects.}

\classificationexample
{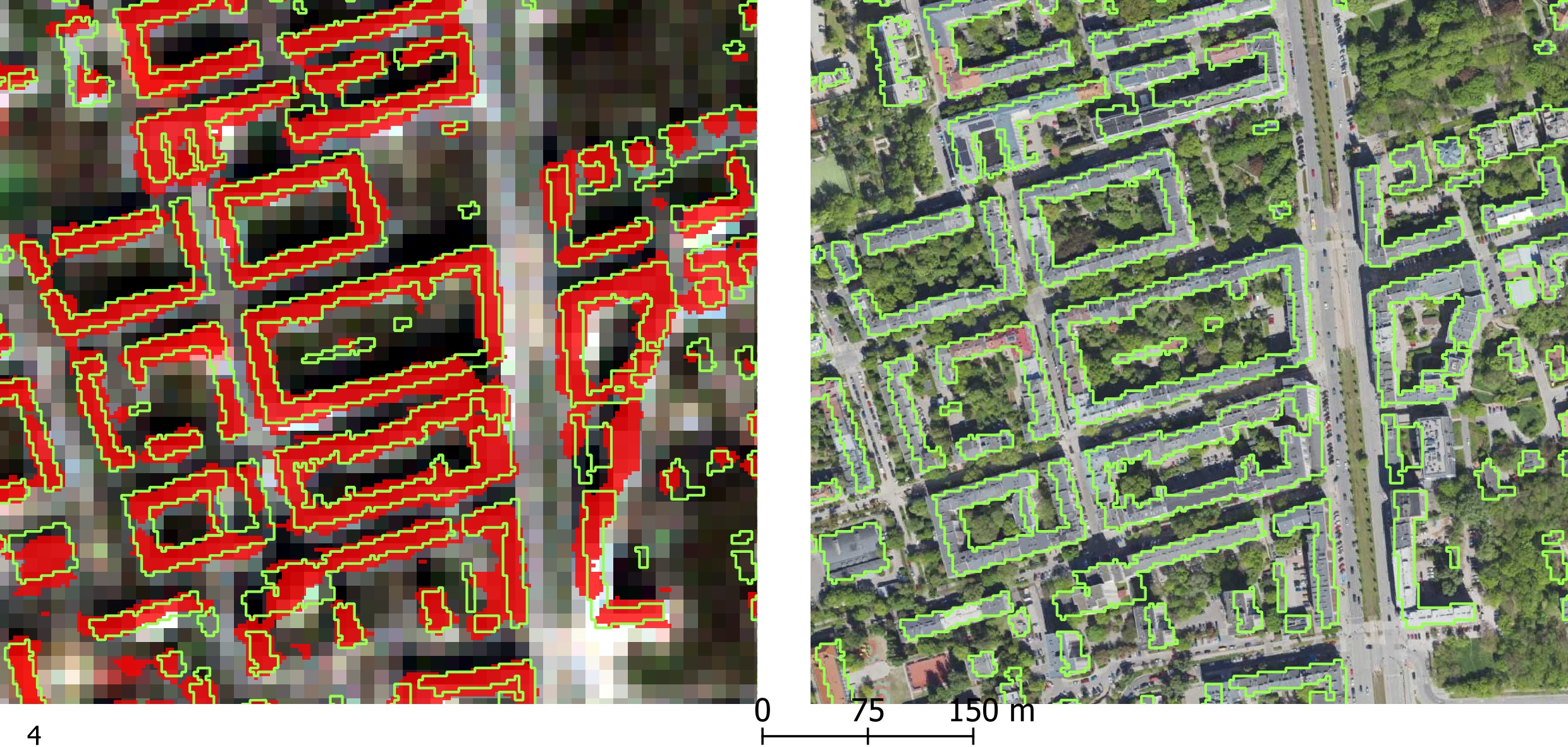}
{High-rise buildings -- accurate detection with seasonal shadow effects}
{This example shows a very good classification of tall, isolated buildings, based on the image from 9 March 2024. Despite the 10 m Sentinel-2 input resolution, the model reconstructs the approximate shape of high-rise structures well. The case also illustrates an important seasonal effect: under low Sun angles and long shadows, classifications may become slightly displaced, often towards the south, or spatially more diffuse. This suggests that shadow geometry can influence localisation accuracy and may be important to consider in application.}

\classificationexample
{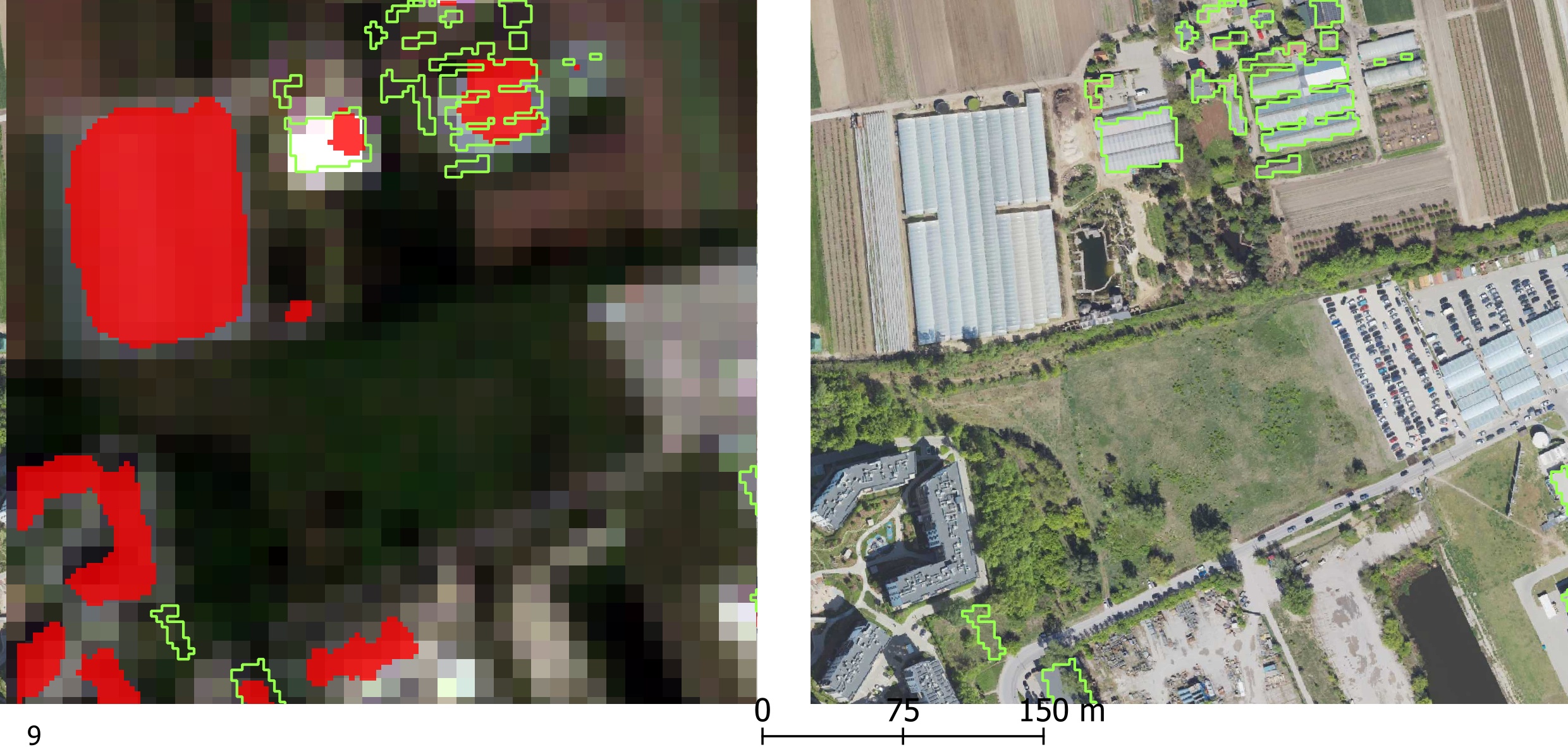}
{Plastic tunnels and greenhouse-like structures -- detections beyond the training set}
{This example shows the detection of plastic tunnels in the image from 12 August 2023. Importantly, the model was not explicitly trained on this type of object, as such structures were not represented in the training set. The result suggests that the method can generalize to related covered constructions, including greenhouses, plastic tunnels, and temporary enclosed sports facilities such as covered tennis courts. This may be practically relevant in the context of taxation, legal compliance, and monitoring of structures that are temporary, newly introduced, or not consistently recorded in official inventories.}

\classificationexample
{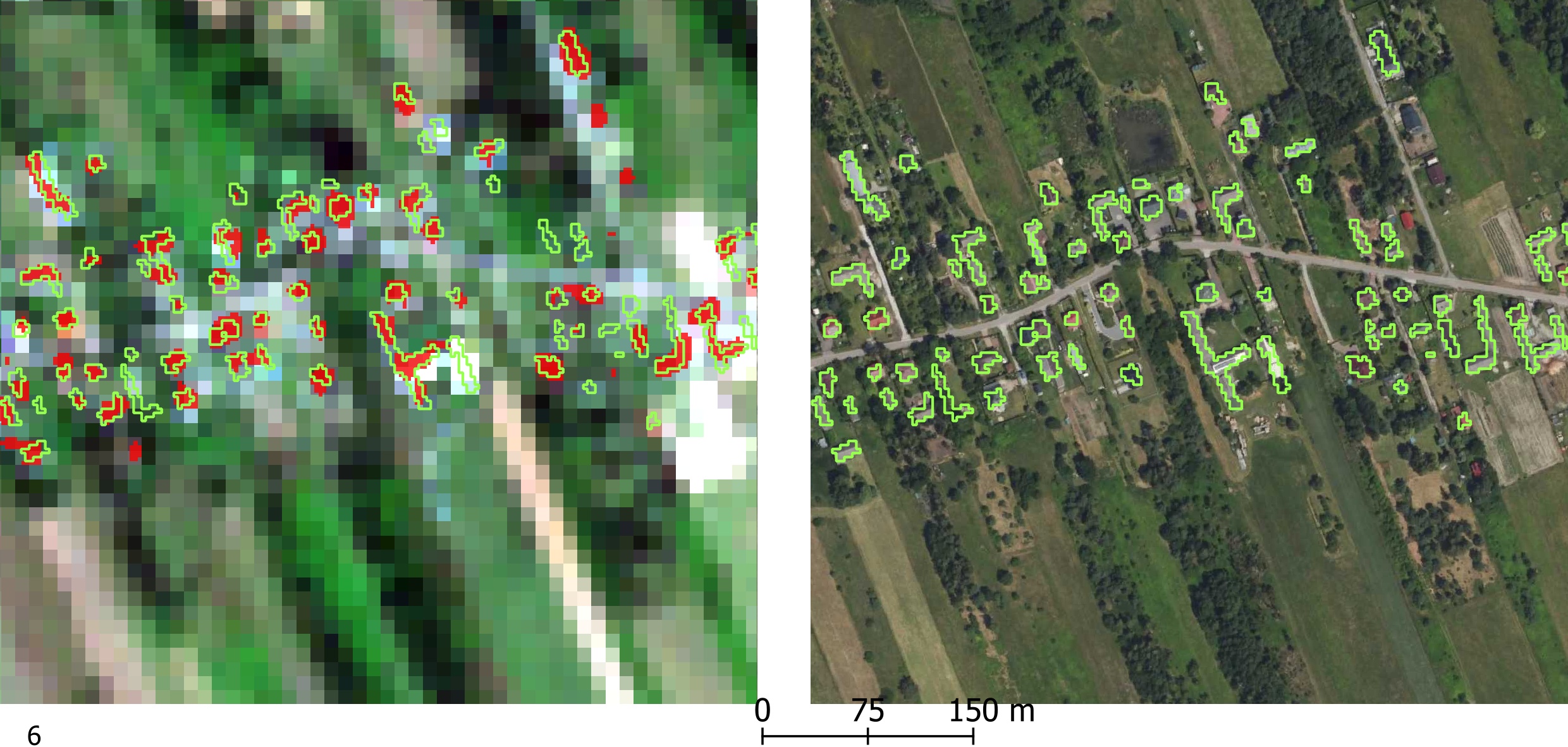}
{Rural small buildings -- promising detection despite resolution limits}
{This example shows a reasonably good classification of small buildings in a rural area, based on the image from 12 August 2023. Given the 10 m source resolution, the result is visually appealing; a task that is particularly difficult for traditional methods. Although the quantitative scores are clearly lower than for compact urban development, the classifications remain convincing from a visual and practical perspective. The main limitation is the expected spatial imprecision: building footprints are often slightly shifted relative to the ground truth or appear somewhat enlarged. Nevertheless, the method appears well suited for identifying newly built structures, detecting illegal or previously unregistered constructions such as sheds or canopies, and monitoring dispersed rural development, where somewhat larger farm buildings can additionally support classification.}

\classificationexample
{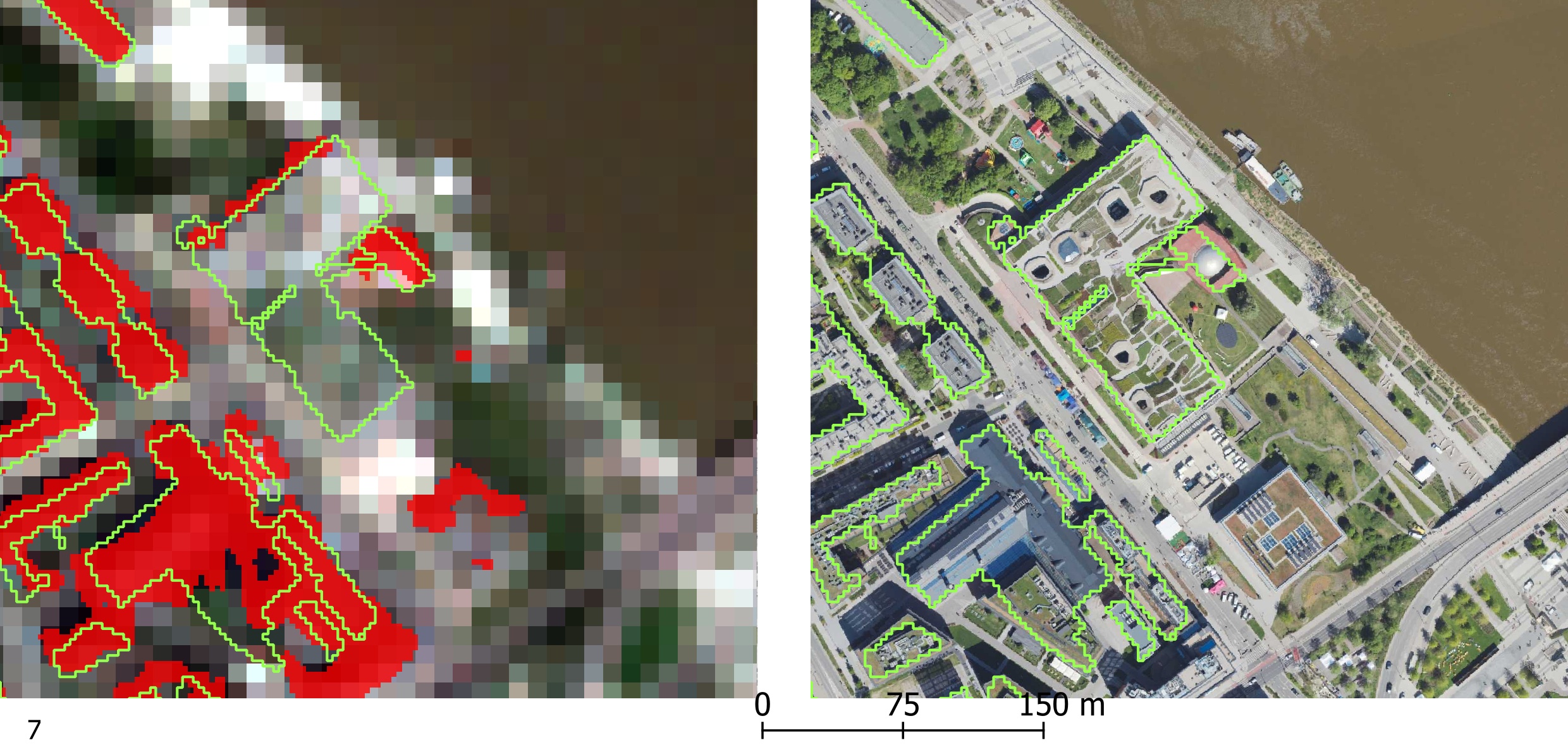}
{Green roofs -- missed detection of buildings with rooftops containing vegatation}
{This example shows a failure to detect buildings with green roofs, illustrated by the Copernicus Science Centre in the image from 12 August 2023. This is an important limitation of the method: vegetated rooftops can be spectrally similar to surrounding green areas and may therefore be missed by the classifier. The issue is particularly relevant for detecting new service, office, or public buildings, where green roof designs are increasingly common.}

\section{Detailed results tables}
\label{app:results-tables}

This appendix presents tabulated values of the evaluation metrics reported in the main text.

\begin{table}[H]
\centering
\caption{IoU, F1 and BA for each measure, obtained within Phase 2 (cross-scene classification) for L2A products, for 8 testing areas (R1 –industrial built-up area, R2 – high density built-up area (city centre), R3 – high density built up area, R4 – medium density built-up area (multi-family blocks), R5 and R6 – suburban, low rise built-up area (single-family buildings), R7 and R8 – dispersed rural built-up area).}
\label{tab:metrics_regions_phase2}
\begin{tabular}{lcccccccc}
\hline
\textbf{ } & \textbf{R1} & \textbf{R2} & \textbf{R3} & \textbf{R4} & \textbf{R5} & \textbf{R6} & \textbf{R7} & \textbf{R8} \\
\hline

\multicolumn{9}{l}{\textbf{Intersection over Union (IoU)}} \\
Mean   & 0.504 & 0.500 & 0.477 & 0.466 & 0.306 & 0.291 & 0.257 & 0.292 \\
Median & 0.532 & 0.519 & 0.508 & 0.493 & 0.339 & 0.324 & 0.280 & 0.328 \\
Std    & 0.092 & 0.063 & 0.096 & 0.091 & 0.095 & 0.093 & 0.090 & 0.103 \\
Min    & 0.078 & 0.226 & 0.085 & 0.084 & 0.013 & 0.003 & 0.000 & 0.009 \\
Max    & 0.587 & 0.558 & 0.560 & 0.538 & 0.398 & 0.386 & 0.354 & 0.400 \\
\hline

\multicolumn{9}{l}{\textbf{F1-score}} \\
Mean   & 0.664 & 0.664 & 0.639 & 0.630 & 0.460 & 0.442 & 0.399 & 0.441 \\
Median & 0.694 & 0.683 & 0.674 & 0.661 & 0.507 & 0.490 & 0.438 & 0.494 \\
Std    & 0.101 & 0.062 & 0.103 & 0.103 & 0.128 & 0.128 & 0.131 & 0.142 \\
Min    & 0.144 & 0.369 & 0.157 & 0.155 & 0.025 & 0.005 & 0.000 & 0.017 \\
Max    & 0.740 & 0.717 & 0.718 & 0.699 & 0.569 & 0.557 & 0.523 & 0.572 \\
\hline

\multicolumn{9}{l}{\textbf{Balanced Accuracy (BA)}} \\
Mean   & 0.805 & 0.814 & 0.811 & 0.826 & 0.722 & 0.722 & 0.700 & 0.741 \\
Median & 0.818 & 0.825 & 0.837 & 0.848 & 0.747 & 0.741 & 0.712 & 0.769 \\
Std    & 0.060 & 0.041 & 0.065 & 0.066 & 0.075 & 0.075 & 0.069 & 0.083 \\
Min    & 0.531 & 0.611 & 0.537 & 0.541 & 0.506 & 0.501 & 0.500 & 0.504 \\
Max    & 0.865 & 0.851 & 0.865 & 0.880 & 0.819 & 0.832 & 0.828 & 0.854 \\
\hline
\end{tabular}
\end{table}

\end{document}